\documentclass[10pt,twocolumn,letterpaper]{article}

\usepackage{iccv}
\usepackage{times}
\usepackage{epsfig}
\usepackage{graphicx}
\usepackage{amsmath}
\usepackage{amssymb}
\usepackage{subfigure}
\usepackage{rotating}
\usepackage{footnote}
\usepackage{array}
\usepackage{rotating}
\usepackage{multirow}
\usepackage{soul}

\usepackage[pagebackref=true,breaklinks=true,letterpaper=true,colorlinks,bookmarks=false]{hyperref}

\iccvfinalcopy  

\def\iccvPaperID{79\vspace{2.1in}} 

\renewcommand{\deg}{\ensuremath{^{\circ}}\xspace}

\begin{document}

\title{A robot that recogconize every objects in a limited-size but unconconstrained space} 
\title{Recogconizing every object perfectly in a limited-size but unconconstrained environment} 
\title{Recogconizing objects perfectly in a limited-size but unconconstrained environment} 
\title{Perfect object recognition in a limited-size but unconconstrained environment} 
\title{Perfect recognition and localization in a limited-size but unconconstrained environment} 
\title{A robot that recognizes every object in a home}

\title{How to Build a Robot that Recognizes Every Object Reliably?}
\title{How far we are from solving computer vision for service robot?}
\title{Realizing Robot Vision Today}
\title{Making Robot Vision Work Today}
\title{Robo. Recognize-All: Making Robot Vision Work Today}
\title{Robo. Know-It-All: Making Robot Vision Work Today}
\title{RoboKnow-All: Making Robot Vision Work Today}
\title{RoboKnowAll: Making Robot Vision Work Today}
\title{Robo. Know-All: Making Robot Vision Work Today}
\title{Robo. Know-It-All: Making Robot Vision Work Today}

\title{Robot In a Room: Making Robot Vision Work Today}

\title{Robot In a Room: Robotic Object Recognition at Human Performance}

\title{Robot In a Room: Almost-Perfect Object Recognition in Closed Environments}

\title{Robot In a Room: Toward Perfect Object Recognition in Closed Environments}


\author{Shuran Song \quad\quad Linguang Zhang \quad\quad  Jianxiong Xiao\\ Princeton University\vspace{2.3in}}

\maketitle

\begin{figure}
\vspace{-2.4in}
\begin{minipage}[b]{1\textwidth}

\includegraphics[width=1\linewidth]{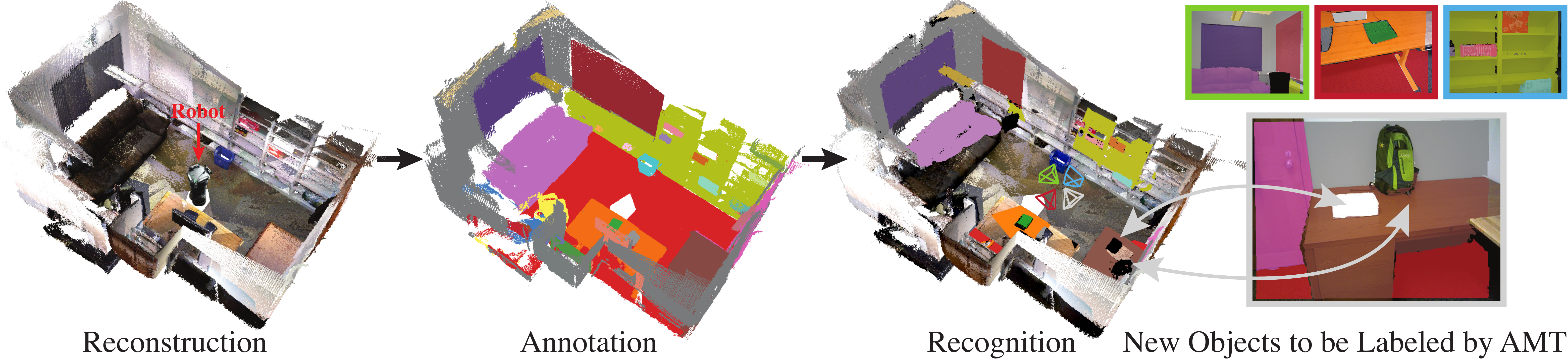}%

\caption{{\bf A robot that can recognize all the objects}.
We propose an extremely robust mechanism to reconstruct a 3D map and use crowd sourcing to collectively annotate all objects.
During testing, the robot localizes its pose, recognizes all seen objects (four images on the right from four RGB-D sensors mounted on the robot), 
and identifies new ones (e.g. the backpack and the box).
In most cases, the robot can recognize autonomously. It can indicate reliably when it fails, and utilize crowd sourcing to fix the problem or to annotate new objects.}
\label{fig:teaser}
\end{minipage}
\end{figure}

\begin{abstract}
\vspace{-2mm}
While general object recognition is still far from being solved, this paper proposes a way for a robot to recognize every object at an almost human-level accuracy. Our key observation is that many robots will stay in a relatively closed environment (e.g. a house or an office). By constraining a robot to stay in a limited territory, we can ensure that the robot has seen most objects before and the speed of introducing a new object is slow. Furthermore, we can build a 3D map of the environment to reliably subtract the background to make recognition easier. We propose extremely robust algorithms to obtain a 3D map and enable humans to collectively annotate objects. During testing time, our algorithm can recognize all objects very reliably, and query humans from crowd sourcing platform if confidence is low or new objects are identified. This paper explains design decisions in building such a system, and constructs a benchmark for extensive evaluation. Experiments suggest that making robot vision appear to be working from an end user's perspective is a reachable goal today, as long as the robot stays in a closed environment. By formulating this task, we hope to lay the foundation of a new direction in vision for robotics. Code and data will be available upon acceptance.

\end{abstract}

\vspace{-5mm}
\section{Introduction}
\vspace{-1mm}

Consider the following grand challenge of computer vision for robotics: prepare a living room for a 10-person party. To enable this task, we need a robot to be able to do planning, grasping, manipulation and reasoning. But before all that, at least the robot needs to recognize the objects, such chairs, sofas, coffee tables, dishes, sodas, etc.

While we have witnessed a dramatically improvement of object recognition in the past few years (e.g. \cite{DPM,AlexNet,RCNN}), we are still far from having a vision system for robots to recognize objects reliably in real-world environments. The success of Internet single-image classification and detection cannot live up to its promise to visual perception in robotics. Therefore, most robot visual perception systems today only focus on recognizing a dozen of carefully pre-scanned object instances with clean background (Figure \ref{fig:robotVision}).

We aim to bridge the gap to enable robots to recognize realistic objects very reliably in their natural settings. Our {\bf key observation} is that many (service) robots will stay in a relatively closed environment, and this greatly constrains the possible objects that it can encounter. For example, a household service robot will stay in a house and never go outside the house \cite{BillGates}. The set of possible objects that the robot can encountered is finite, and the speed of introducing a new object into a typical house is also limited. A robot doesn't have to recognize all the objects in the world to be fully capable in such a closed environment. This paper proposes to exploit this critical constraint to build a robot vision system that can reliably recognize objects in a realistic environment at an almost human-level accuracy.

Although this paper focuses on robot vision, the same idea also applies to many other movable devices that are only used in closed environments, such as head-mounted displays (e.g. Oculus Rift or Microsoft HoloLens) that people typically only use in their own living rooms.

\begin{figure}[t]
\vspace{-3mm}
\centering

\includegraphics[width=1\linewidth]{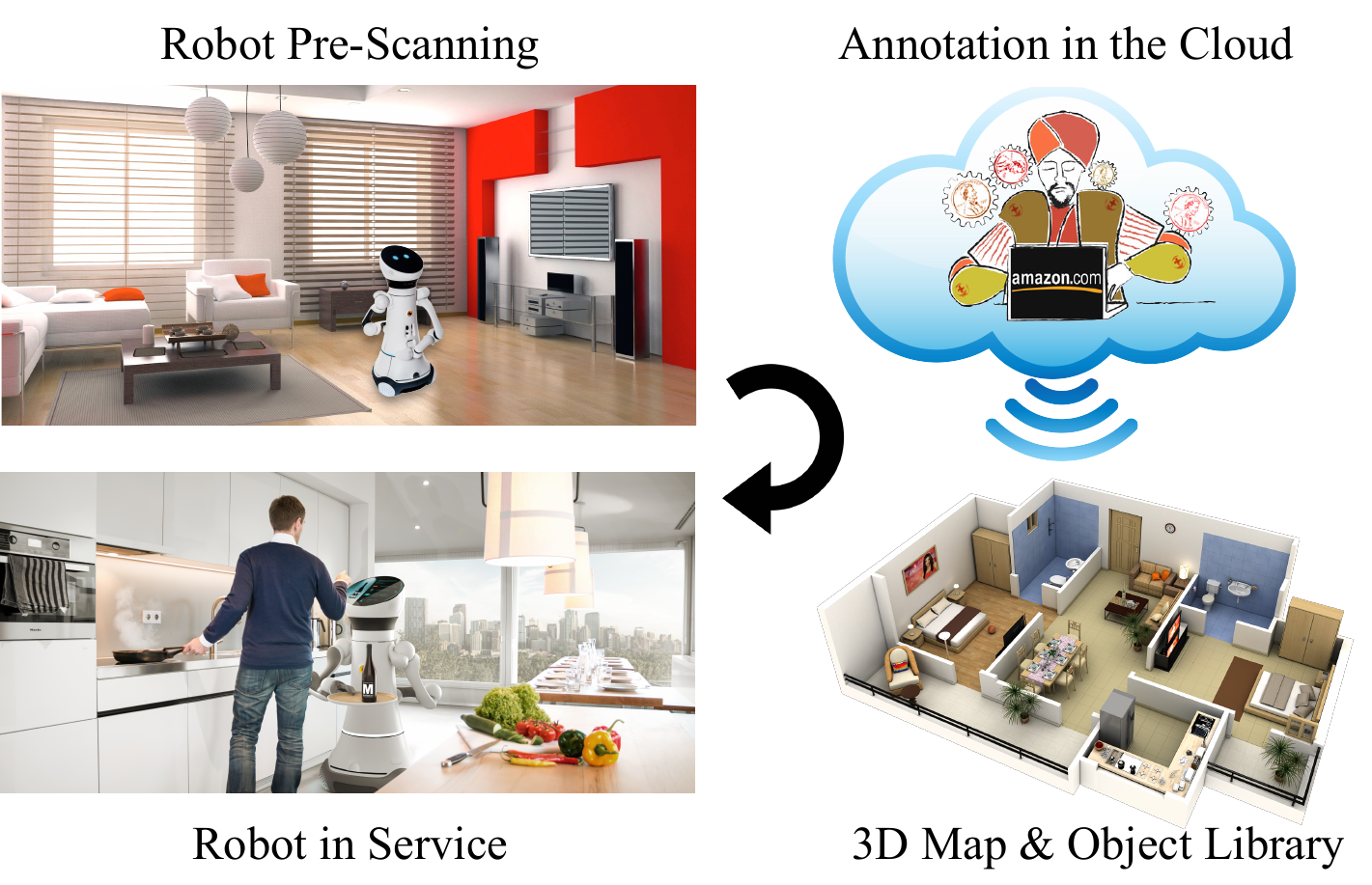}

\vspace{-1mm}
\caption{{\bf User scenario.} A robot pre-scanned a house. All data is uploaded to the cloud to be annotated by human workers on crowd sourcing platform,
to obtain a object library and a 3D map. After that, the robot can recognize all the objects and provide service.}
\label{fig:scenario}

\vspace{-4mm}
\end{figure}

\subsection{User scenario}

A user buys a robot, brings it home and turns it on. And the journey of its life begins, as illustrated in Figure \ref{fig:scenario}. In the first few hours, the robot does nothing but wanders around the house, scanning the environment by taking videos to reconstruct a 3D map of the place. 
The robot picks
some frames from the video to cover the space, and asks human workers in the cloud to label all the objects in these frames,
using online crowd-sourcing platforms (e.g. Amazon Mechanical Turk). 
For a typically apartment, this costs about \$100 and takes about an hour. 
The robot uses the annotation to build a 3D object library and it is ready for service.


Now the robot can navigate within the apartment and locate itself, relying on the main structure of the rooms that are not movable (e.g. no home remodeling). Given the location, the limit set of possible objects, and the knowledge of their appearance specific to this apartment, the robot can recognize the objects almost perfectly. Of course, the robot will break miserably if it leaves the specific apartment, but it can avoid going outside the known territory. If there is unrecognizable change (e.g. new object), it will reconnect to crowd-sourcing platform again and ask humans to annotate the frame (e.g. with a monthly subscription fee of \$10). In such a way, a robot can recognize all the objects in the house perfectly to support its missions.

\subsection{Challenges and our solutions}

To build such a system, there are several technical challenges that we have to overcome (Figure \ref{fig:teaser}). First, we need an extremely robust way to reconstruct a 3D map for any given environment. In this paper, we propose two coupled mechanisms to make this happen: a panoramic RGB-D camera array and a special robot path to ensure significant view redundancy for loop closing. Second, we need a way to enable multiple humans to collectively and efficiently label all objects in a place at the instance level. We propose an algorithm to select frames to maximize coverage, and a robust way to associate instances in 3D. Third, we need a reliable way to recognize objects during testing time. We propose to use the 3D pose of the robot to retrieve the unmovable background structure of the environment (e.g. floor, walls, and ceiling) in order to reliably subtract the background, which significantly reduces the difficulty of recognizing foreground objects. For each of the foreground movable objects, we build a mixture of 3D models merging from multiple views to increase view invariance. Finally, we need a reliable way to fill in holes of old objects when the occlusion situation changes to avoid requesting human annotation too frequently, but at the same time to avoid overly propagate to uncertain areas or new objects. To this end, we propose to integrate object color, normal distribution, shape bounding box, spatial continuity, 
to eliminate impossible object labels for reliable propagation.

\begin{figure}[t]
\vspace{-2.2mm}
\centering

\includegraphics[width=1\linewidth]{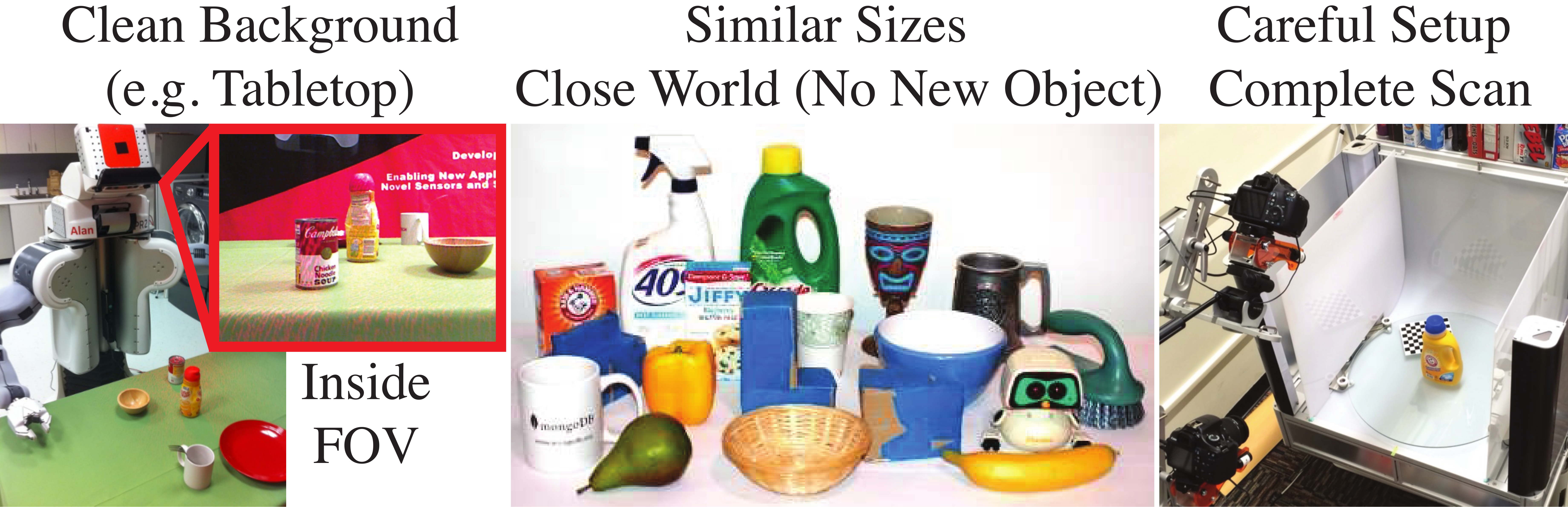}

\vspace{-0.5mm}
\caption{{\bf Current object recognition for robotics} mostly focuses on 
recognizing objects that are completely inside the field of view (FOV) on a clean background.
The objects are from a very limited set with similar sizes suitable for those algorithms.
No new objects are allowed, and all objects are completely scanned using a careful setup.
We advocate to recognize objects in their natural scenes without change to the environments. (image source: \cite{GloverBingham,BigBIRD})}

\label{fig:robotVision}

\vspace{-6mm}
\end{figure}

Besides these technical contributions, there are several new concepts as well. First, this project provides a way to bring the success of robotic object recognition from tabletops to real-world scenes. Instead of constraining the background to be very artificial, we recognize objects in realistic environments but we are still able to use background subtraction to ease recognition. Second, unlike Internet images, there is ample domain knowledge we can use to reduce the difficulty of perception in robotics applications. This task puts computer vision algorithms to a real test. By integrating all useful cues, it is a great exercise to let us clearly know how far we are from making robot vision work. Third, our user scenario may potentially incubate a new business model for robot perception, by leveraging automatic recognition and online crowd sourcing. Fourth, our formulation of the user scenario opens up a new research direction in vision for robotics. By constructing a realistic offline benchmark directly related to the goal, we can test algorithms without a robot but reliably reflexing the true performance in real world. Last, we have built a working system with all the components. Experiments shows that we can already make object recognition working most of time without any human involvement to enable autonomous operation.

\subsection{Related works}

There is a vast literature on object recognition from 2D, 3D, RGB-D, video in computer vision and robotics.
For category-level recognition,
the state-of-the-art object detectors are \cite{RCNN,DPM,ESVM}, and \cite{SlidingShapes,DepthRCNN} for RGB-D images.
\cite{Xiaofeng,NYU,GuptaRGBDseg,bo2013unsupervised,lai2012detection} are popular semantic segmentation systems.
However, category-level recognition is still far from human performance.
For instance-level recognition,
well-known approaches include \cite{Lana3Dobject,Goggle3Dlandmark}.
For RGB-D images,
the state-of-the-arts \cite{AbbeelBestPaper,GloverBingham}
focus on recognizing table-top objects on a clean artificial background,
with object models carefully pre-scanned from all view angles \cite{uwDataset,BigBIRD} (Figure \ref{fig:robotVision}).
Our approach is built on top of these success, 
extending them 
to realistic scenes.

Our 3D mapping is related to RGB-D reconstruction \cite{KinectFusion,UWmapping,Kintinuous,niessner2013hashing,Vladlen2015,SUN3D} and localization \cite{shotton2013scene,PushmeetRelocalization}.
Our algorithm is closest to the RGB-D Structure from Motion (SfM) from \cite{SUN3D}.
We extend their algorithm to utilize four RGB-D sensors and encode the camera height as a hard constraint.
We design a special trajectory to control the robot to move in a way with significant redundancy to favor loop closing
to ensure good 3D reconstructions.
Toward semantics, 
there are several seminal works on combining 3D mapping and object recognition on RGB-D scans 
\cite{finman2013toward,SLAMpp,fioraio2013towards,whelan3d,herbst2014toward}.
There are also several seminal works in image domain as well \cite{SemanticSFM,Bao_CVPR2011_SSFM,GeometricPhrases,LeeLayout}.

Our object annotation is built on the success of many predecessors \cite{LabelMe,VATIC,VideoSegmentation}.
The most related one is the SUN3D annotator for RGB-D videos \cite{SUN3D},
which uses object annotation to correct reconstruction errors by object-based loop closing.
Because our 3D reconstruction is much better,
we label individual frames to reduce the labeling difficulty.
We also propose a novel 3D instance association algorithm to link object instances in 3D automatically so that multiple humans can collectively label the same sequence.

Overall, perhaps the most relevant work is {Google's self-driving car} \cite{GoogleCar}.
The great success of the car 
relies on the pre-annotated maps that contains 3D scan of all the streets that the car can drive to. 
The maps are precisely annotated by humans for all stationary objects, such as the location of all traffic signs and lights. 
During driving, based on the car's current location,
the data is looked up to serve as virtual infrastructures, 
and it already gets to know a great deal about the environment.
For example, without even looking, the car already knows
where a traffic light is, and just needs to classify the color of the light.
In such a way, 
the very difficult street scene understanding task is transformed into a much easier instance-level task, 
and human-level recognition accuracy is achieved.
This paper proposes to build {an indoor version of Google's car by utilizing the {\bf power of pre-annotated data}}.
We show that it is a reachable goal today for a general-purpose robot vision system to recognize every object in a closed environment.

\begin{figure}[t]
\vspace{-3mm}
\centering

\includegraphics[width=1\linewidth]{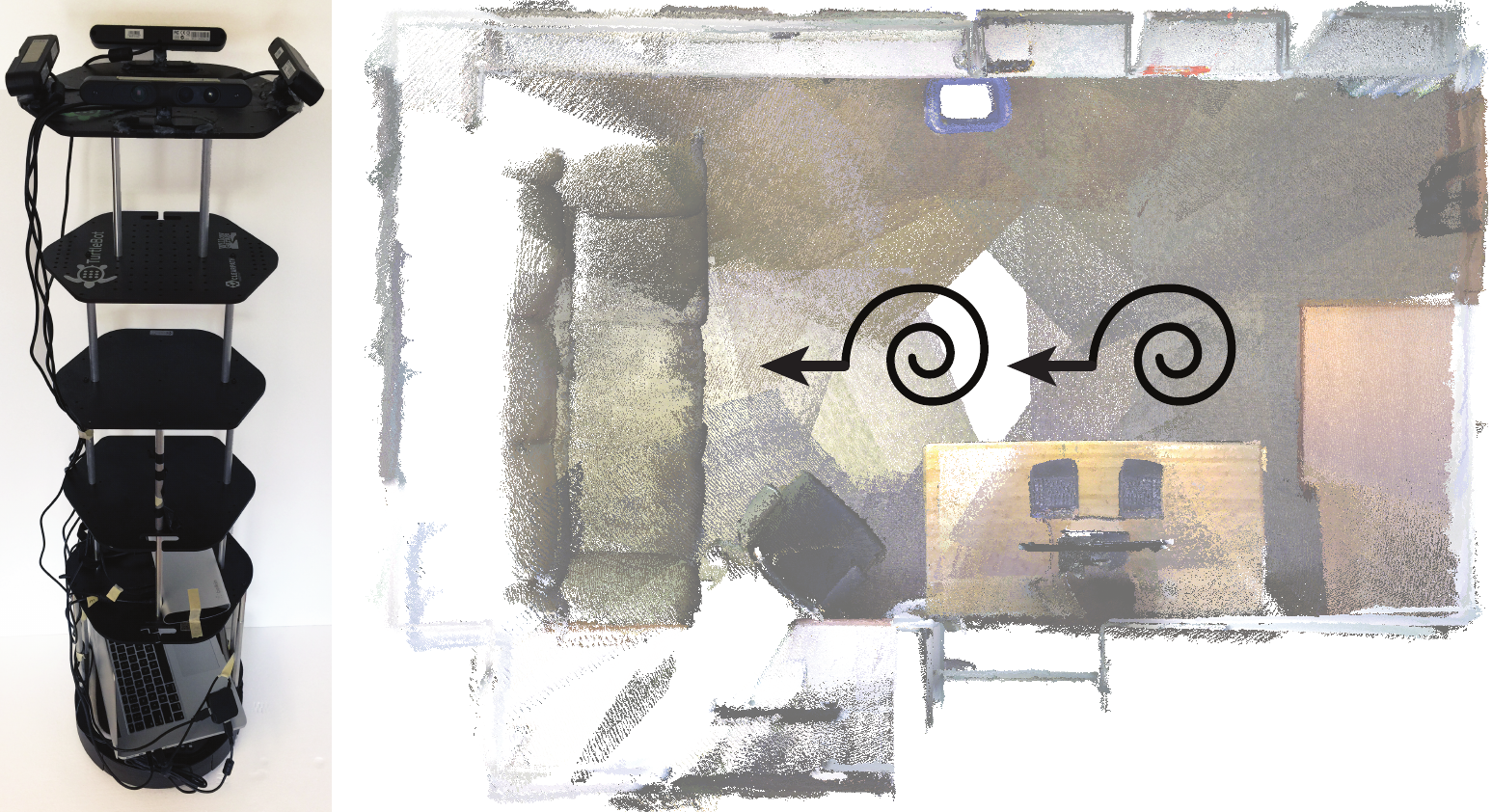}

\caption{{\bf Robot and path.} On the left we show our robot prototype equipped with four RGB-D sensors at the top close to a typical human height.
On the right we illustrate the robot path during training phase,
where it rotates two circles at each location and moves 0.3 meter to the next location.}
\label{fig:RobCtrl}

\vspace{-4mm}
\end{figure}

\section{Formulation}

\vspace{-2mm}

To make it concrete,
we formulate the goal as the following task.
There are two phases: training and testing.
During {\bf training} phase, 
RGB-D videos are captured when the robot moves around in the environment that it is supposed to spend its life in.
None of the objects is allowed to move during this phase.
The 3D map of the environment is reconstructed, and a small number of frames are picked to be labelled by human workers on Amazon Mechanical Turk.
During {\bf testing} phase,
for each of the RGB-D frames captured by the robot, 
the task is to estimate the 6D camera pose for the frame,
and recognize the object categories and 6D object poses for all the objects presented in the images.
Objects may be at different locations between training and testing, and among different testing images.
Some objects may disappear and some new objects may appear.
The goal is to produce a very high quality recognition result with a reliable confidence indicator,
to enable autonomous operation in most cases.
The algorithm should also be able to tell whether it is necessary to bother human annotators, if the recognition lacks confidence or when some new object is identified.
Note that our focus is on the vision component, and other components of the robot decide its trajectory during testing phase, depend on its missions.

\vspace{-2mm}

\section{Training phase}
\label{sec:reconstruction}

\vspace{-1mm}

During training, we desire a robust way to reconstruct the 3D map that the robot will spend its life in.
The success of whole system depends on the 3D map,
and it is crucial to be able to have a very robust algorithm for any environment.
To this end,
we propose two major mechanisms to make reconstruction become much easier.
First,
we mount four RGB-D sensors on top of the robot 
to have a panoramic RGB-D input.
For example, if the robot can measure the distance to the four walls of a typical rectangular room,
it is much easier to localize the robot to obtain a good 3D camera pose.
Because the robot has a fixed height, we can also enforce the camera height as a hard constraint.
Second,
we let the robot to move in a way that it sees a lot of duplicate views to enable loop closing.
When bundle adjustment is overly constrained with many correspondences, 
3D reconstruction will have no choice but to converge to the right solution.

\begin{figure}[t]
\vspace{-3mm}
\centering

\includegraphics[width=1\linewidth]{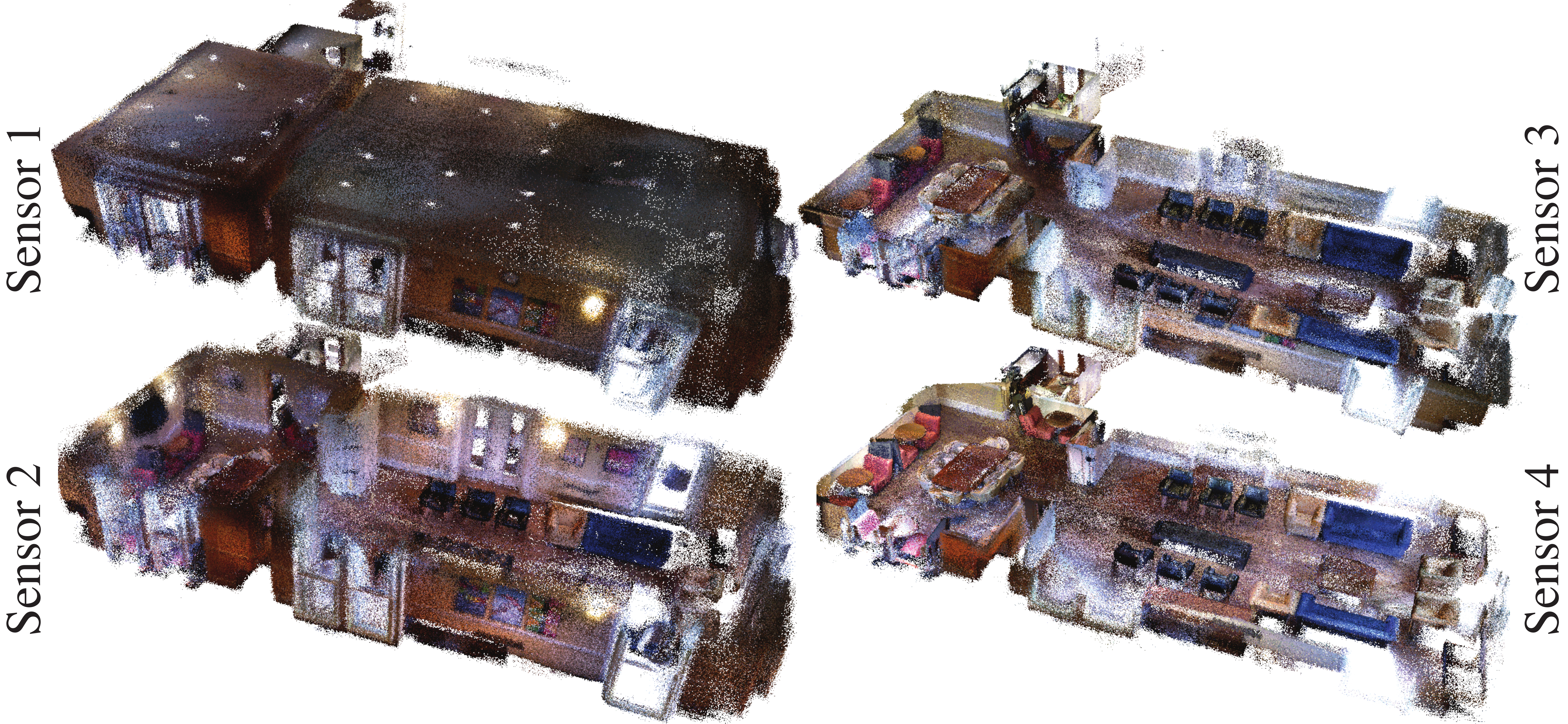}

\caption{{\bf 3D reconstruction from diffferent sensors.} Sensor 1 sees the ceiling, while sensor 4 sees the floor. }
\label{fig:LayerRec}

\vspace{-4mm}
\end{figure}

\subsection{RGB-D camera array}

We use an off-the-shelf TurtleBot 2 (Figure \ref{fig:RobCtrl}) and a Macbook Pro.
The laptop controls the robot wheels and collects data from four ASUS Xtion Live Pro sensors.
We use four sensors to stream four RGB-D videos at $640\times480$ 30fps, reaching the maximal USB bandwidth of the laptop.
We mount them on a shelf that is 1.4 meter tall,
because most indoor spaces are designed for human heights.
All the components of our robot can be purchased off-the-shelf from the Internet, and cost about \$2000 (excl. laptop).

We configure the four cameras to have a panoramic field of view.
They are arranged at 0\deg, 90\deg, 180\deg and 270\deg horizontally,
and 5\deg up, 10\deg down,  25\deg down and 40\deg down vertically, covering 90\deg vertical field of views.
Figure \ref{fig:LayerRec} contains point clouds reconstructed from the cameras with different tilt angles.
We didn't use a checkerboard to calibrate of the transformation between cameras,
because our cameras have no view overlapping at all.
Instead, we use the hardware setting to initialize the calibration and adjust it during reconstruction.

\subsection{Scanning path} 
To make object recognition more robust, we desire to densely capture objects from as many different views as the robot can possibly navigate to.
Also, to produce a good reconstruction,
we control the robot to move in a way that enforces significant view overlapping to ensure ample loop-closing correspondences.
As shown in Figure \ref{fig:RobCtrl},
we let the robot move 0.3 meter, stop to rotate two circles, and then move to the next location.
Rotating two circles lets every camera see exactly the same view twice.
In this way, every frame of the RGB-D videos will have many keypoint correspondences with other frames.

\subsection{Panoramic SfM}
\label{sec:RGBDsfm}

Similar with the state-of-the-art RGB-D SfM \cite{SUN3D},
we use a frame-based approach with 3D keypoint-based bundle adjustment.
For each RGB-D frame, we detect SIFT keypoints on the color images,
and use depth map to obtain the 3D coordinates for the keypoints.
For a pair of RGB-D images, we use RANSAC with 3 points to estimate the 6D rotation and translation between the two images and obtain the inlier correspondences.
We use this pairwise alignment routine
to align every two consecutive frames from the video captured by the same camera to obtain initial camera poses.
We use the initial calibration among the cameras to align the 3D reconstructed from the four cameras as the initial poses for bundle adjustment.
To obtain more correspondences,
we compute bag-of-word for each image,
and choose pairs of frames from any camera with high dot product values.
For each pair, we use the pairwise alignment routine to obtain inlier correspondences to add into the bundle adjustment.

\begin{figure}[t]
\vspace{-3mm}

\centering
\includegraphics[width=1\linewidth]{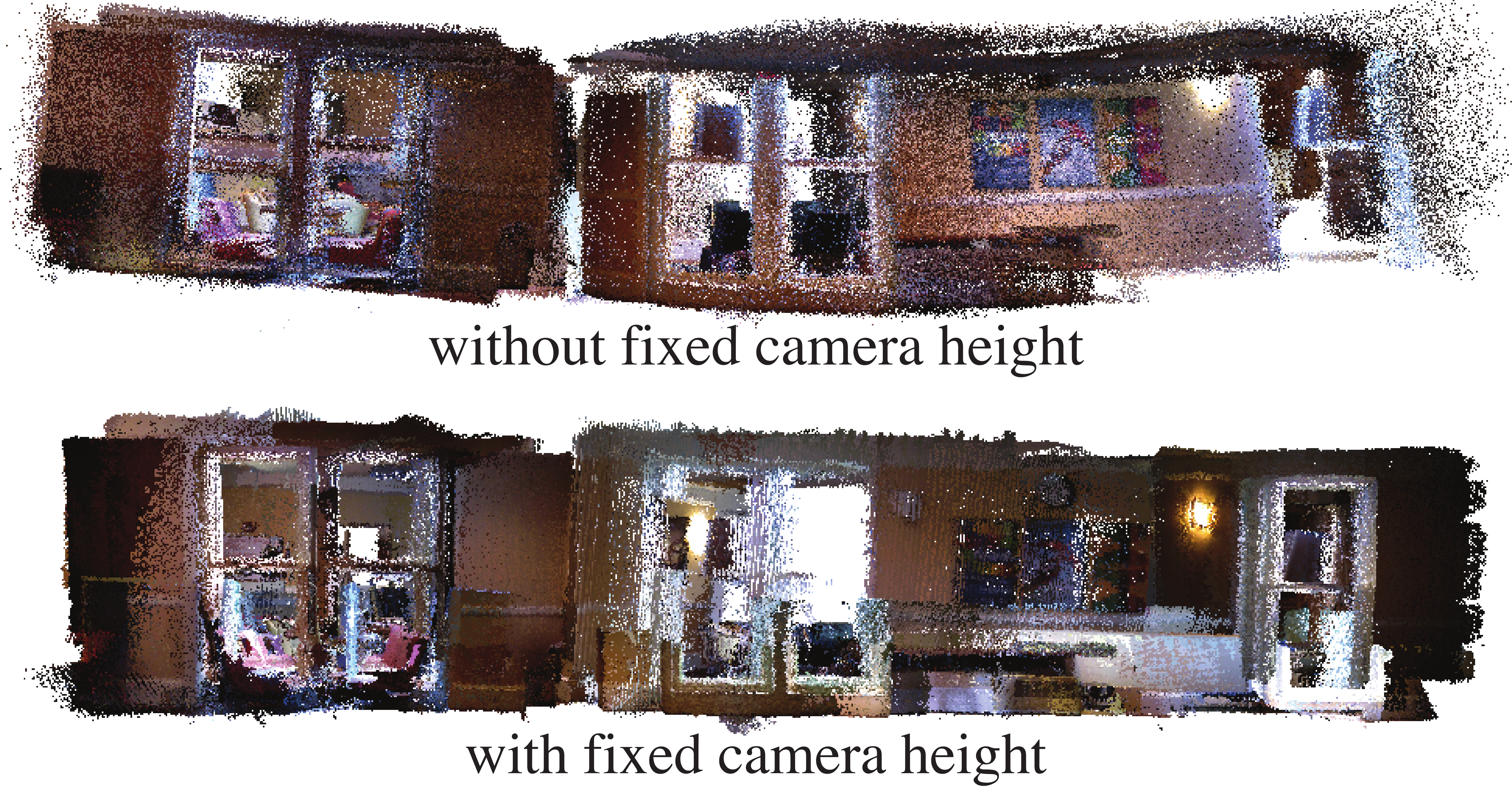}

\vspace{-1mm}
\caption{{\bf Camera height as hard constraints in bundle adjustment.}
We can see that by constraining the camera height, the floor gets flat and the reconstruction gets better.}
\label{fig:CamHeight}

\vspace{-4mm}
\end{figure}

For the $n$-th frame,
we model the robot base by its $(x_n,z_n)$ location and rotation angle $\theta_n$, 
with $y_n=0$ because the robot has a fix height ($\vec{y}$ is the gravity direction).
We model the extrinsics of each camera as a 6D rotation $\mathbf{R}_c$ and translation $\mathbf{t}_c$ w.r.t. the robot base. 
For a 3D point $\mathbf{X}$ from the $c$-th camera of $n$-th frame, its world coordinate is
$ R_{\vec{y}}(\theta_n) ( \mathbf{R}_c \mathbf{X} + \mathbf{t}_c ) + [x_n,0,z_n] ^ T$, 
where $ R_{\vec{y}}(\cdot)$ is a $3\times3$ rotation matrix for rotating around $\vec{y}$ axis.
For bundle adjustment \cite{ceres-solver},
we model these extrinsic parameters $\{x_n,z_n,\theta_n\}$ and the camera calibration $\{\mathbf{R}_c,\mathbf{t}_c \}$ as variables, 
with the default intrinsic parameters from OpenNI as constants.
We did try to keep the camera calibration  $\{\mathbf{R}_c,\mathbf{t}_c \}$ as constants during bundle adjustment.
But the reconstruction results are always better when we model them as variables. 

\begin{figure}[t]
\vspace{-3mm}
\centering

\includegraphics[width=1\linewidth]{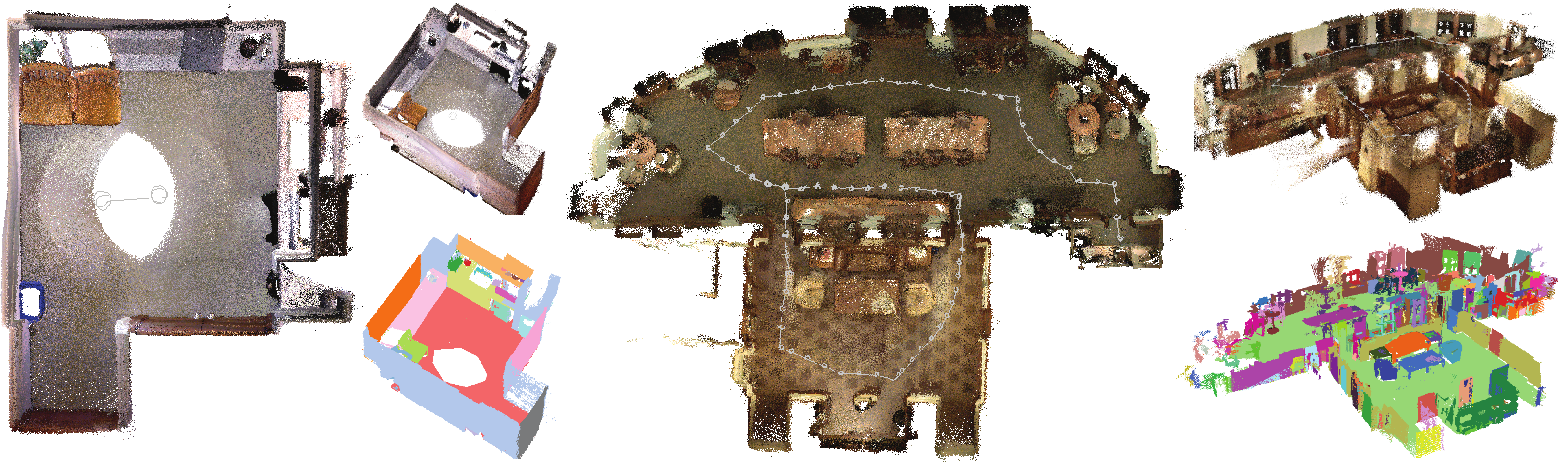}

\vspace{-1mm}
\caption{{\bf Reconstruction \& annotation.} The kitchen in on the left, and the tea area is on the right. Semantics is encoded as colors.}
\label{fig:3DreconResult}
\end{figure}

To speed up reconstruction of long sequences,
we break down a sequence into segments of 1,000 frames.
We reconstruct each segment independently. 
Then, we link these segments together by aligning the consecutive frames among them.
For each frame in a segment, we look for a frame with significant view overlapping from another segment, 
and use the pairwise alignment routine to add more correspondences.
Finally, we bundle adjust the whole reconstruction again.
Figure \ref{fig:CamHeight} compares the reconstruction without and with camera height as a hard constraint.
It takes about 2 hours to reconstruct this sequence using 10 CPU cores.
We tried various environment and found that our system is extremely robust.
Figure \ref{fig:3DreconResult} shows more results.

%

\begin{figure}[t]
\vspace{-3mm}
\centering

\includegraphics[width=0.5\linewidth]{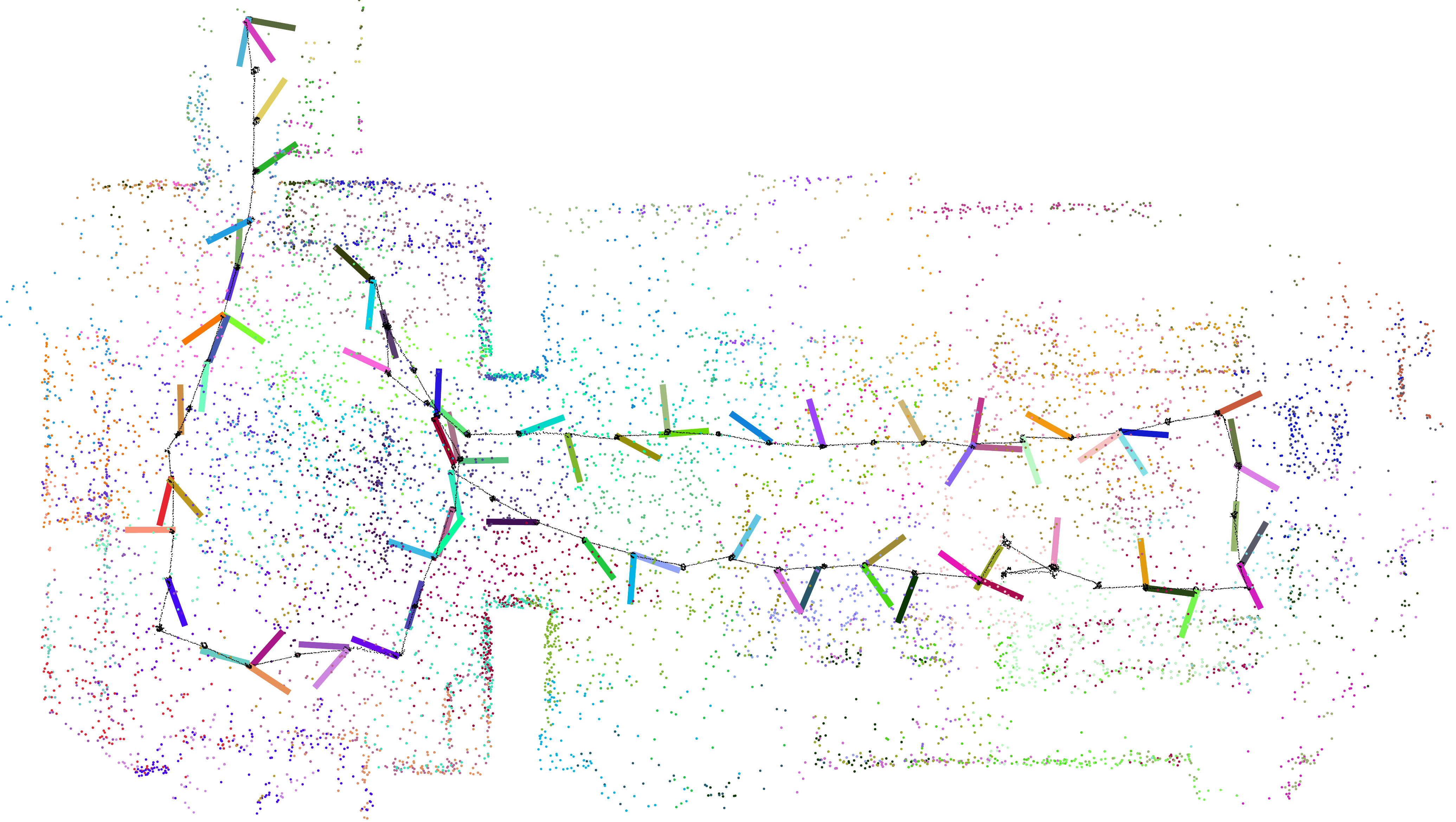}~\includegraphics[width=0.5\linewidth]{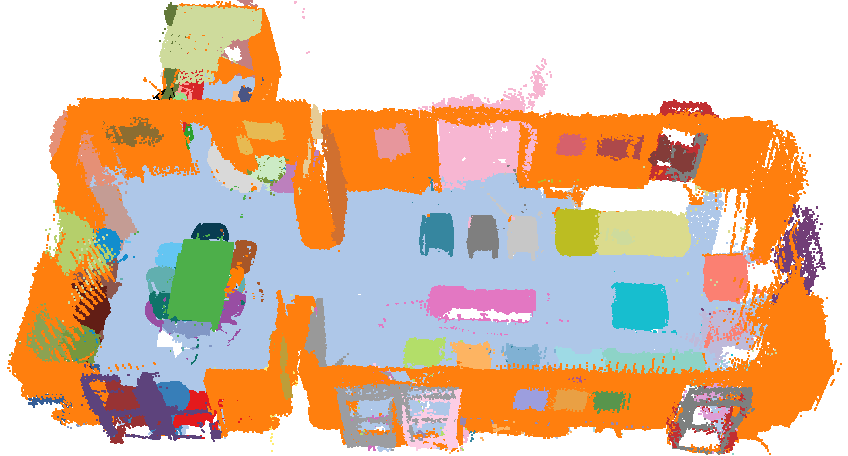}

\caption{{\bf Annotation.}
Left shows the selected images for annotation from the top view.
The polyline shows the camera trajectory, and the colored sticks are the camera direction of selected views.
The 3D points are the voxel centers colored based by their membership to selected views. Right is the annotated 3D point could,
color coded by object instances after automatic association.}
\label{fig:ViewSelection}

\vspace{-4mm}
\end{figure}

\vspace{-1mm}

\subsection{Collectively annotation}
\label{sec:labeling}

It is critical to annotate all the objects in the space at a high quality for reliable recognition.
SUN3D \cite{SUN3D} provides a way to label RGB-D videos.
But because their reconstruction from hand-held captured videos has significant drifting,
they can only rely on local stitching.
In our case, given the high quality reconstruction,
our goal is to minimize human efforts by embracing 3D information as much as possible.

Our algorithm first picks a small number of frames to cover the whole space.
Then, we use Amazon Mechanical Turk to annotate all the objects in each frame by polygons and object categories.
Finally, our algorithm figures out 
the association of polygons to the same instance of objects.

\vspace{-5mm}\paragraph{View selection.}
To minimize the human effort, 
we design an algorithm to choose a minimal number of views to achieve 100\% coverage.
First, we represent the 3D space by voxel grids of $0.01^3$ meter$^3$
and obtain a list of surface voxels that contains more than five 3D points.
We maintain a counter on each voxel to count how many times this voxel has been seen so far.
In each iteration of the algorithm,
we pick one frame from the remaining ones that maximizes the number of voxels with count transition from $0\hspace{-1mm}\rightarrow\hspace{-1mm}1$.
When a frame is picked, we increase the count for each voxel on the depth map of the picked frame.
This greedy selection ends when no new frame can be picked to further increase the coverage.
Figure \ref{fig:ViewSelection} shows an example of view selection.

\begin{figure}[t]
\vspace{-3mm}
\centering

\scriptsize

\begin{flushright}

\begin{sideways}~~~desk\end{sideways}~~%
\includegraphics[width=0.13\linewidth]{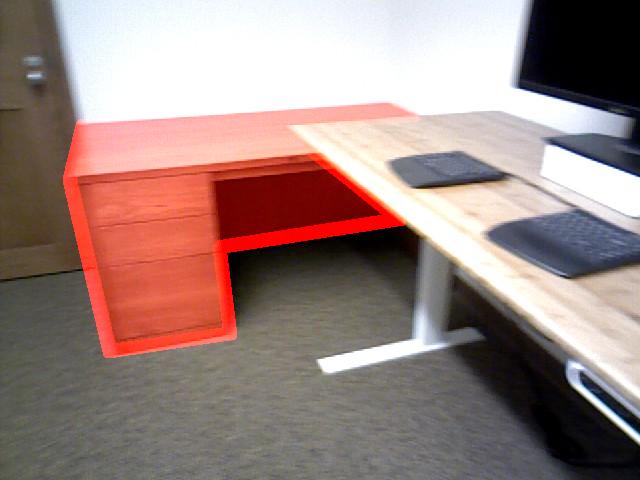}~%
\includegraphics[width=0.13\linewidth]{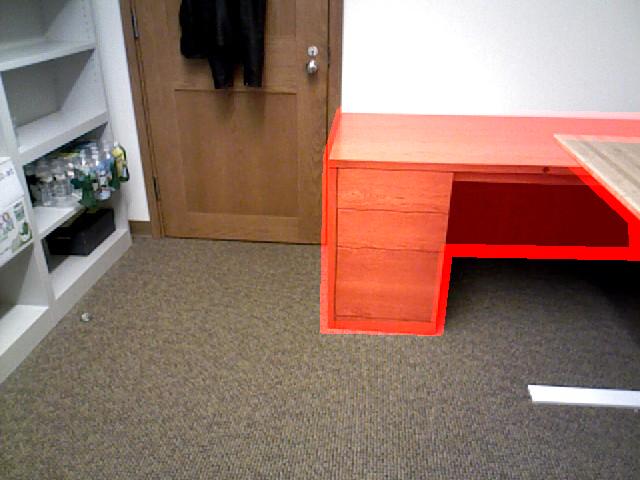}~%
\includegraphics[width=0.13\linewidth]{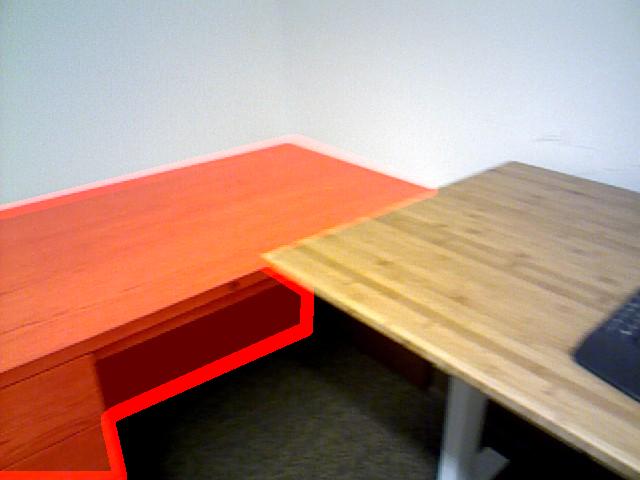}~%
\includegraphics[width=0.13\linewidth]{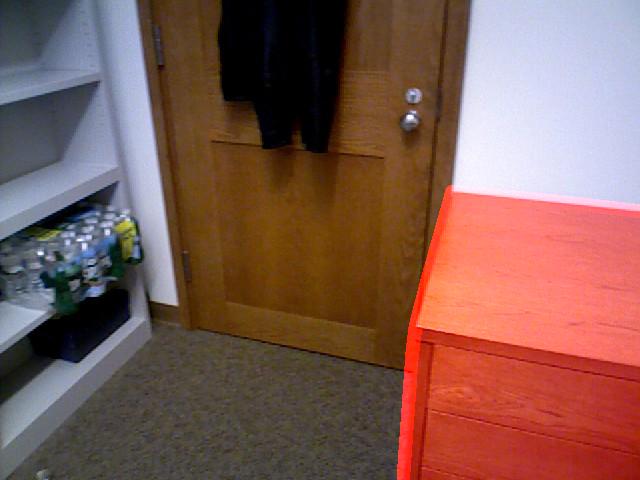}~%
\includegraphics[width=0.13\linewidth]{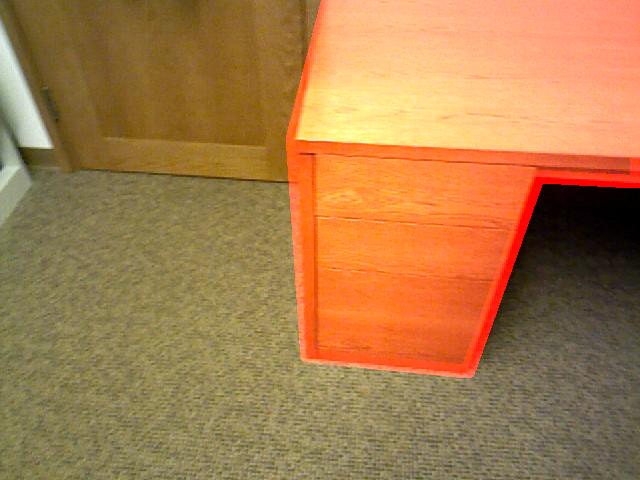}~%
\includegraphics[width=0.13\linewidth]{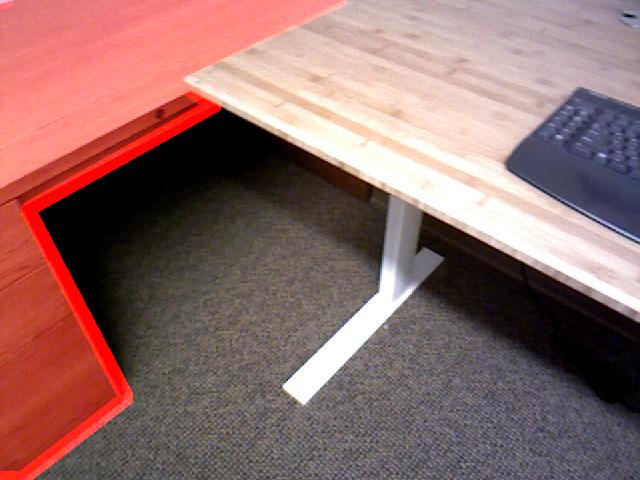}~%
\includegraphics[width=0.13\linewidth]{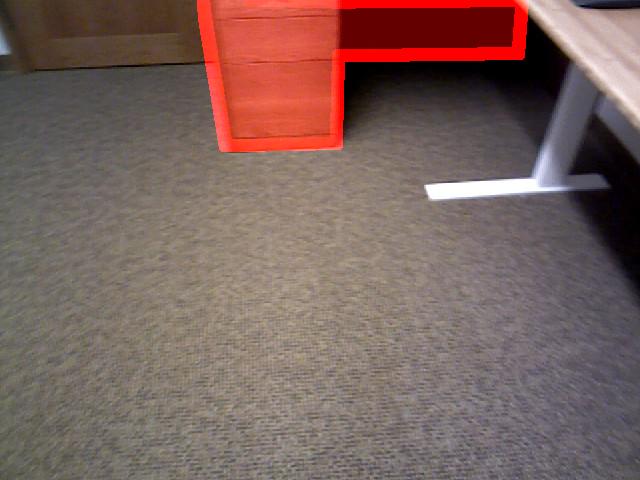}

\vspace{0.7mm}

\begin{sideways}~recycle\end{sideways}~%
\includegraphics[width=0.13\linewidth]{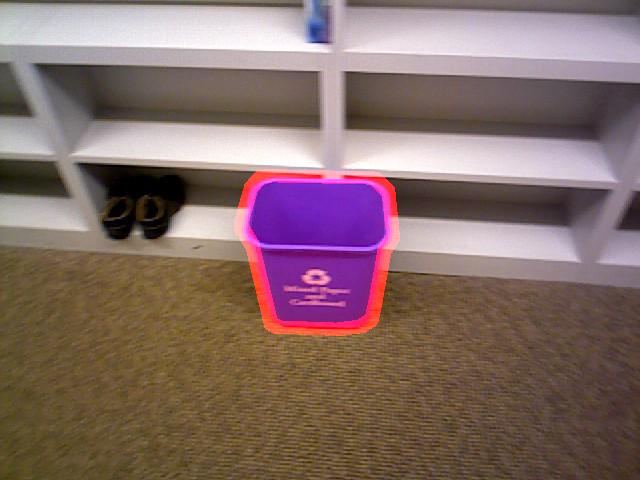}~%
\includegraphics[width=0.13\linewidth]{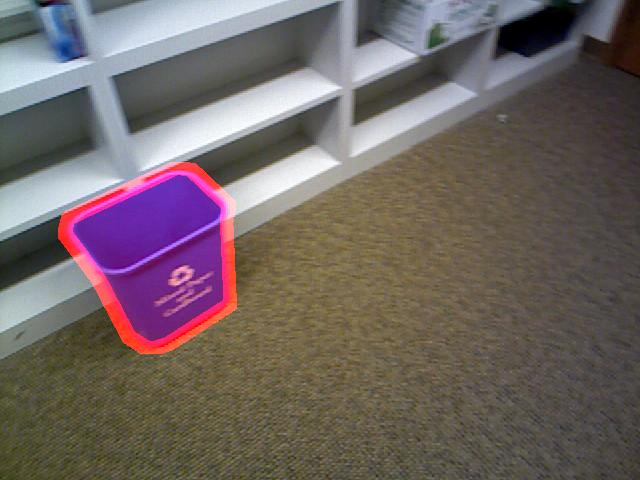}~%
\includegraphics[width=0.13\linewidth]{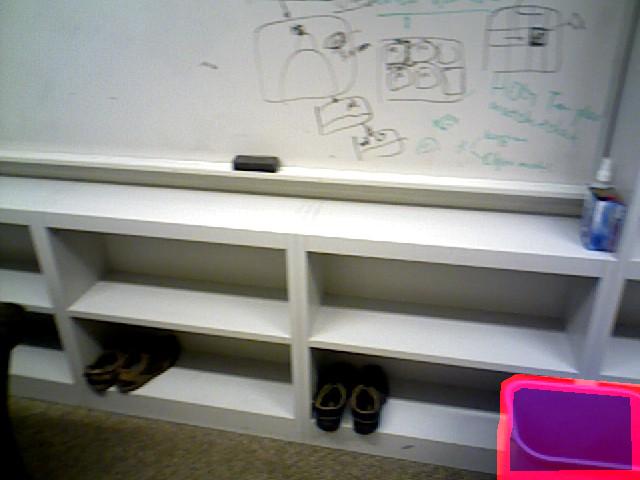}~%
\includegraphics[width=0.13\linewidth]{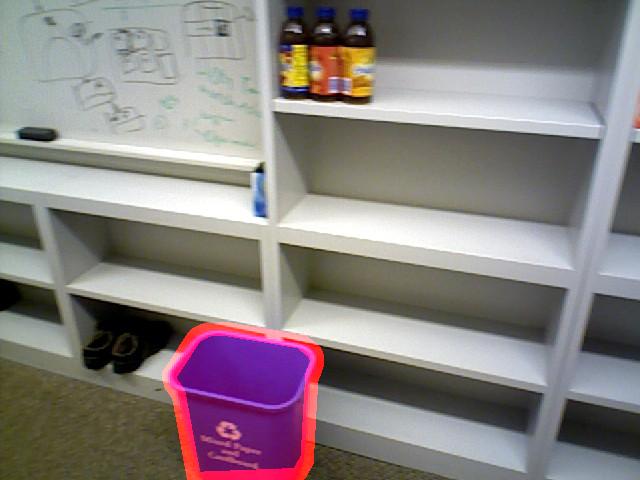}~%
\includegraphics[width=0.13\linewidth]{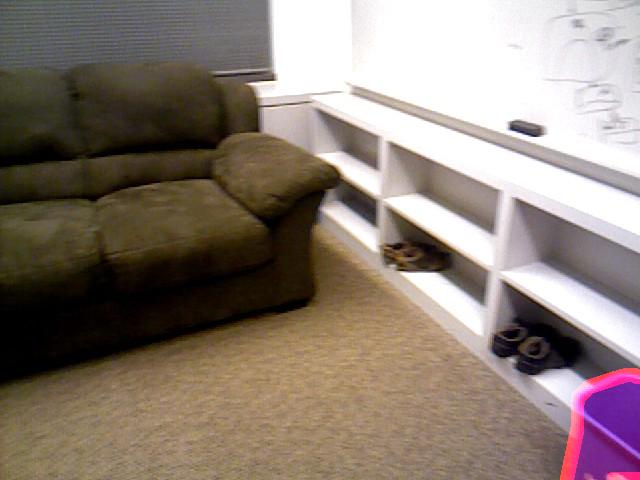}~%
\includegraphics[width=0.13\linewidth]{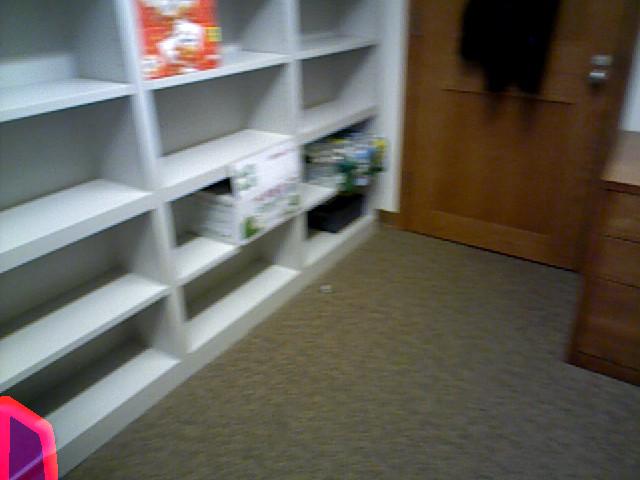}~%
\includegraphics[width=0.13\linewidth]{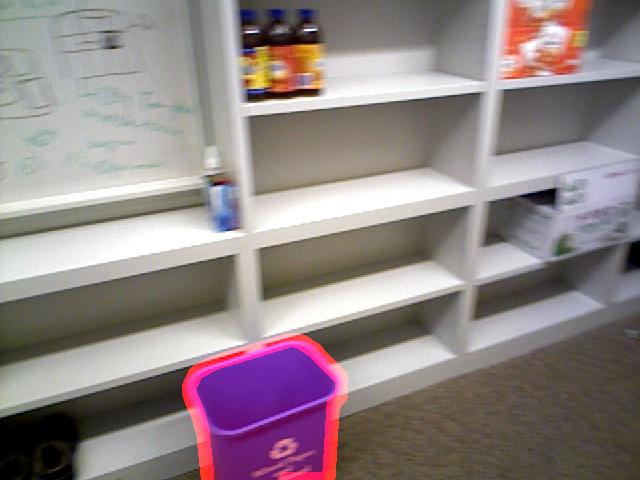}

\vspace{0.7mm}

\begin{sideways}~~~chair\end{sideways}~~%
\includegraphics[width=0.13\linewidth]{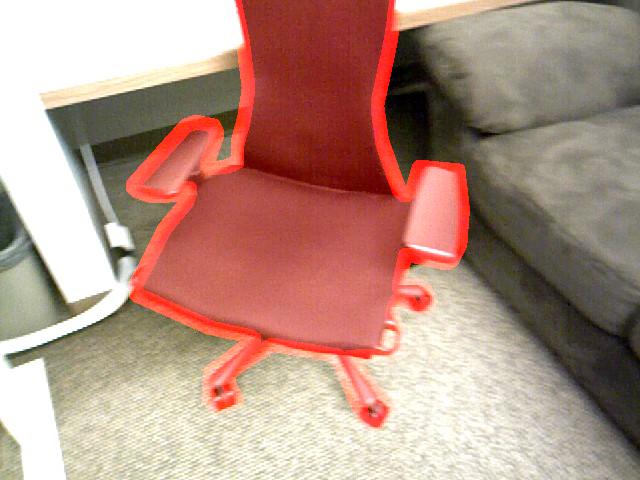}~%
\includegraphics[width=0.13\linewidth]{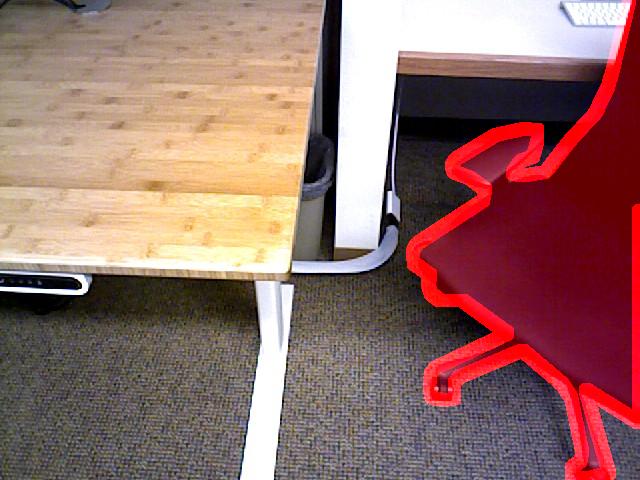}~%
\includegraphics[width=0.13\linewidth]{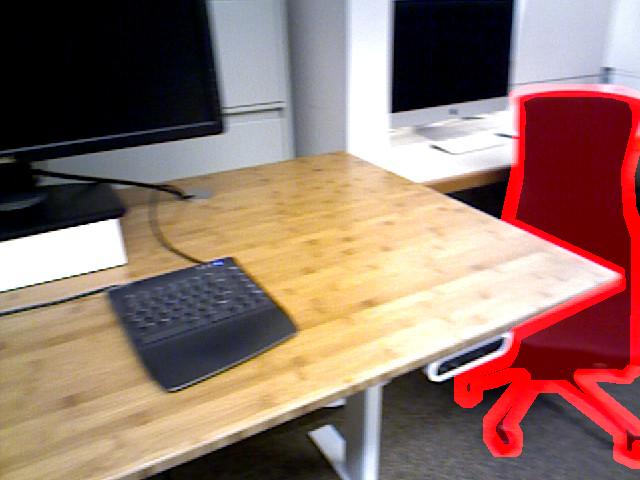}~%
\includegraphics[width=0.13\linewidth]{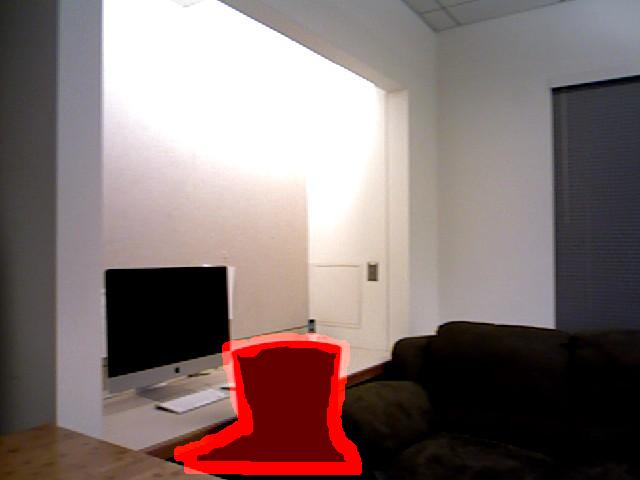}~%
\includegraphics[width=0.13\linewidth]{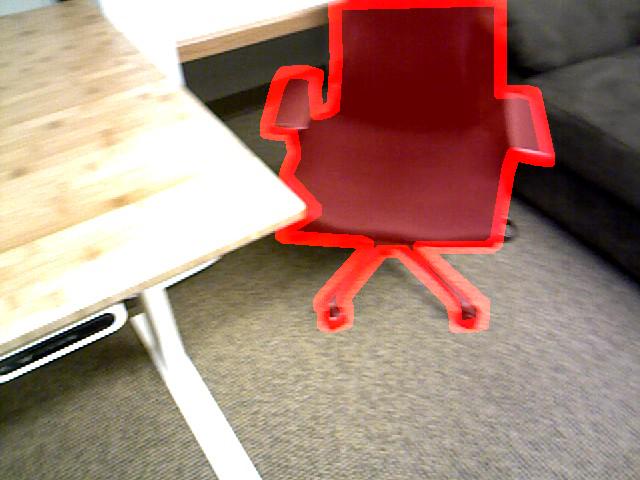}~%
\includegraphics[width=0.13\linewidth]{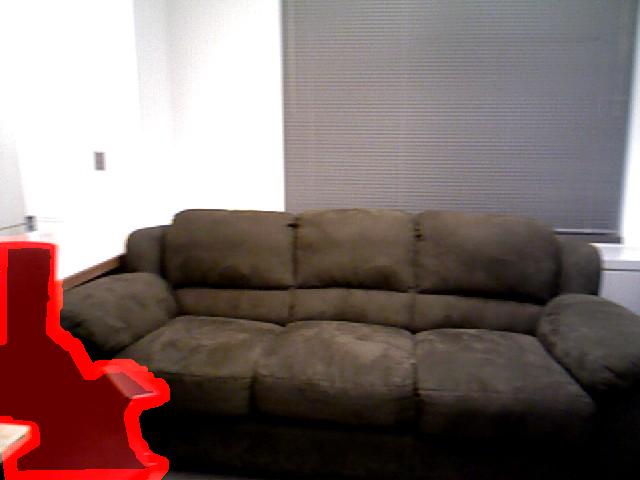}~%
\includegraphics[width=0.13\linewidth]{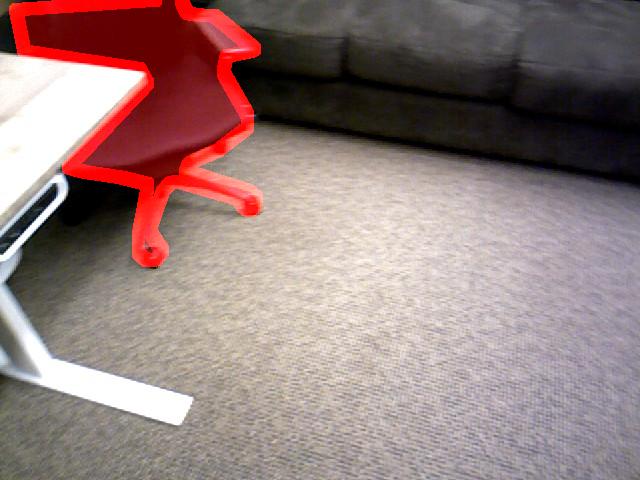}

\vspace{0.4mm}

\begin{sideways}~monitor\end{sideways}~~%
\includegraphics[width=0.13\linewidth]{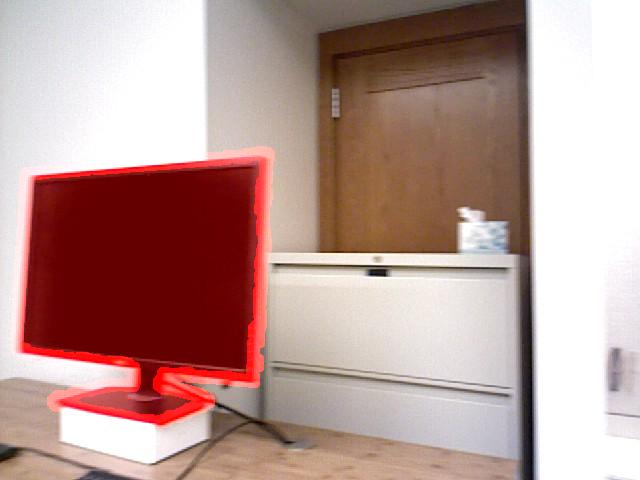}~%
\includegraphics[width=0.13\linewidth]{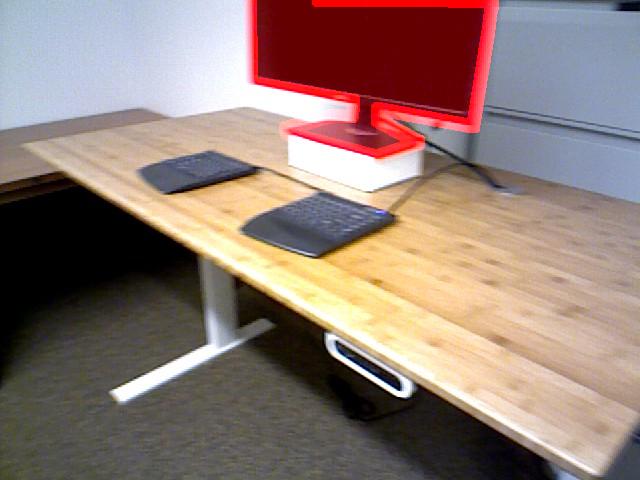}~%
\includegraphics[width=0.13\linewidth]{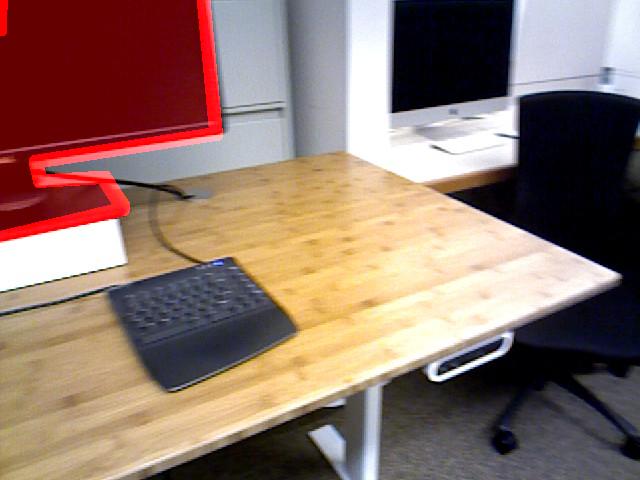}~%
\includegraphics[width=0.13\linewidth]{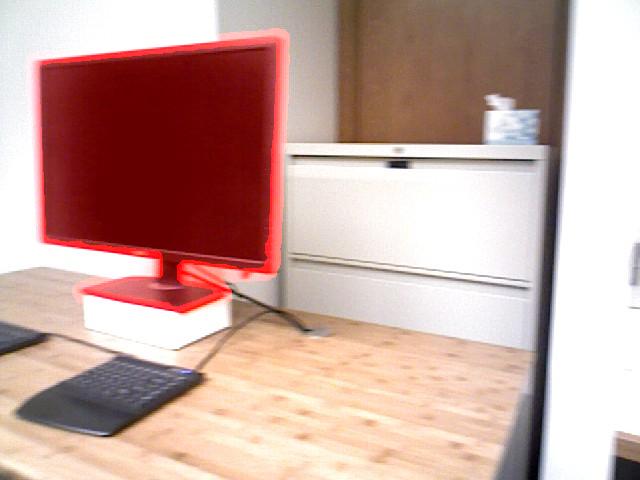}~%
\includegraphics[width=0.13\linewidth]{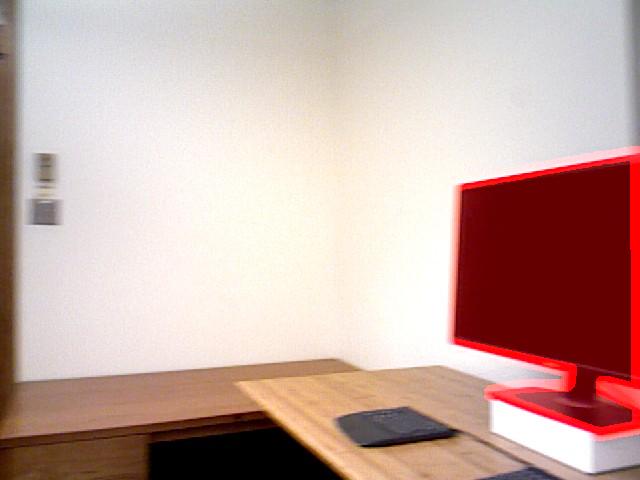}~%
\includegraphics[width=0.13\linewidth]{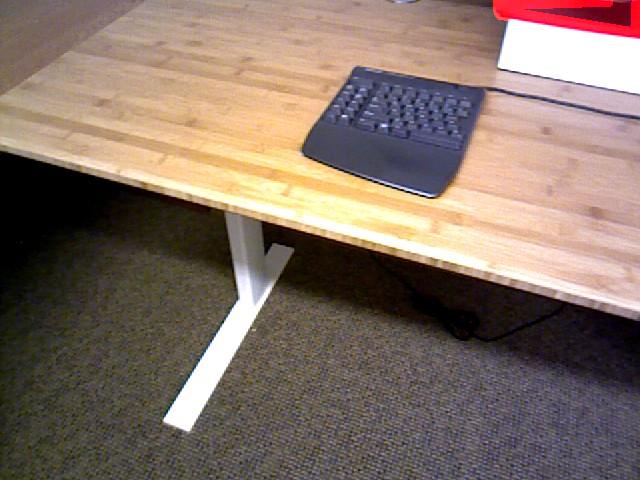}~%
\includegraphics[width=0.13\linewidth]{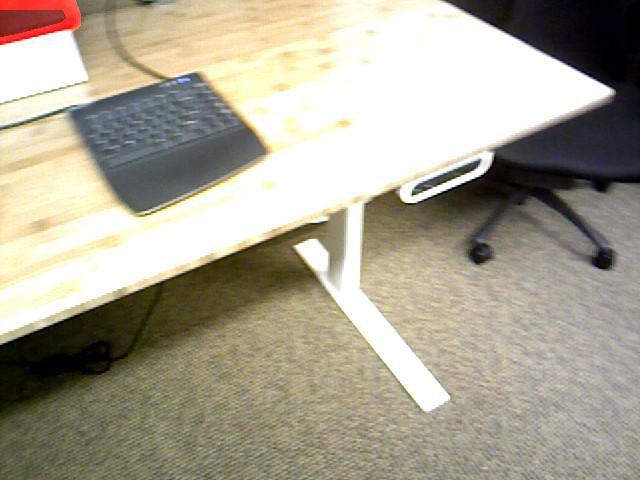}

\vspace{-0.4mm}

\begin{sideways}keyboard\end{sideways}~%
\includegraphics[width=0.13\linewidth]{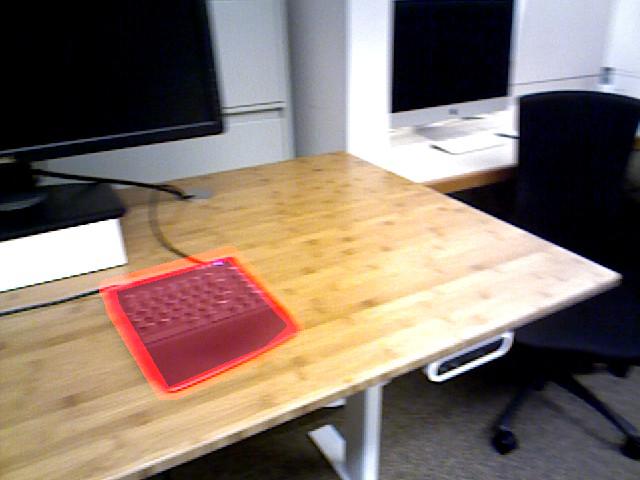}~%
\includegraphics[width=0.13\linewidth]{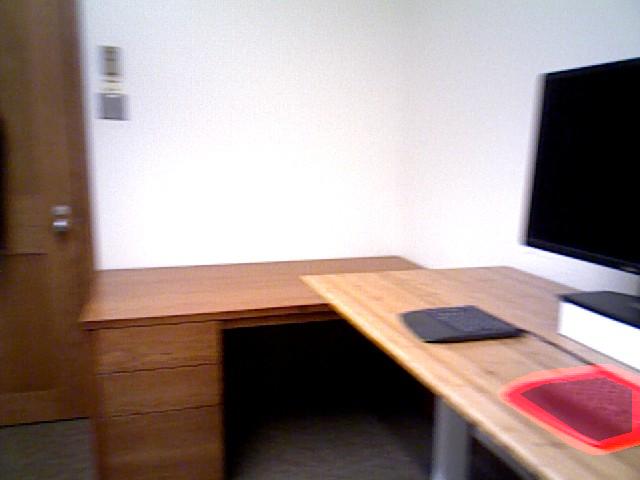}~%
\includegraphics[width=0.13\linewidth]{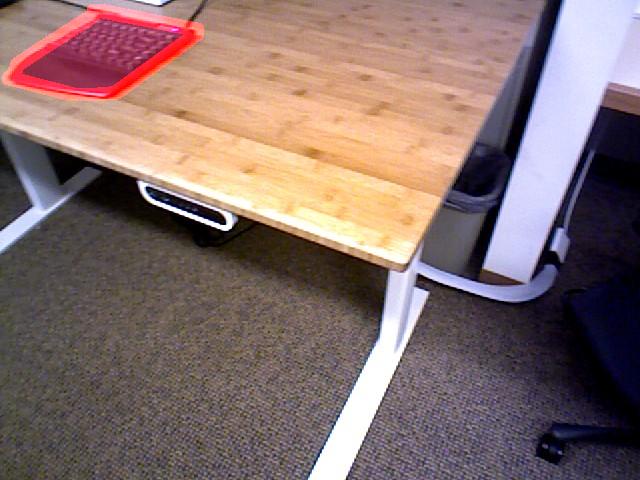}~%
\includegraphics[width=0.13\linewidth]{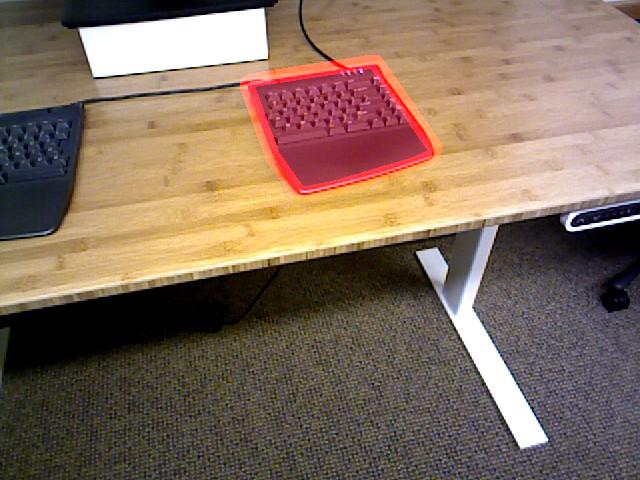}~%
\includegraphics[width=0.13\linewidth]{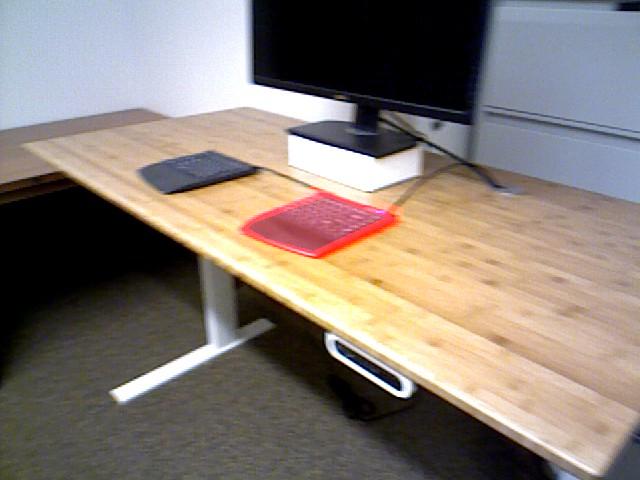}~%
\includegraphics[width=0.13\linewidth]{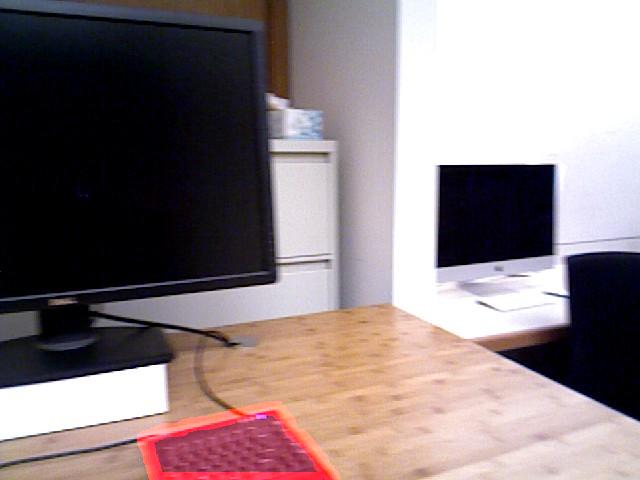}~%
\includegraphics[width=0.13\linewidth]{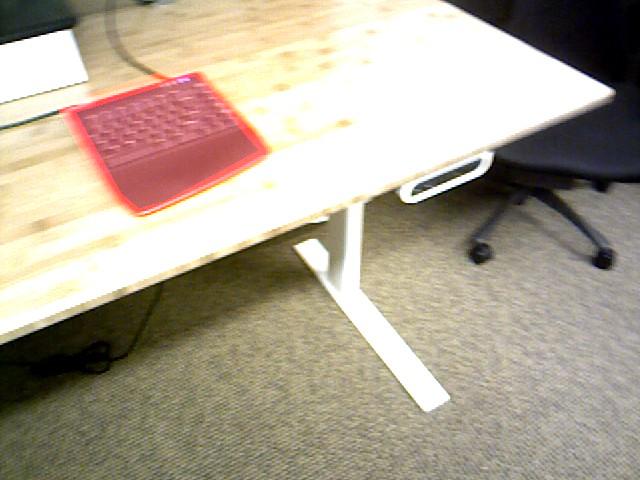}

\end{flushright}

\vspace{-4mm}
\caption{{\bf Instance association.}
Each row shows seven views of the same object automatically associated by our algorithm.
The focus objects are highlighted in red.}
\label{fig:association}

\vspace{-3mm}
\end{figure}

\begin{figure}[t]
\centering

\includegraphics[width=1\linewidth]{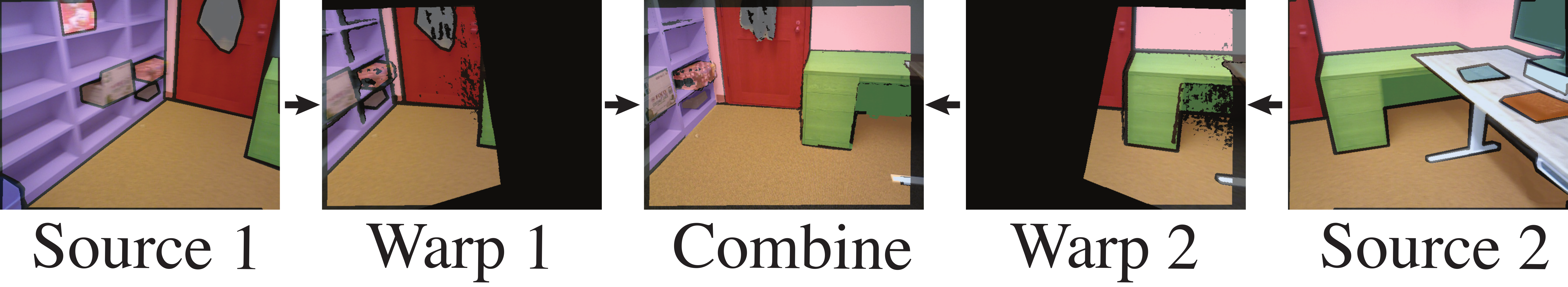}

\vspace{-1.5mm}
\caption{{\bf Label propagation to other view.}}
\label{fig:propagate2view}

\vspace{-4mm}
\end{figure}

\begin{figure*}[t]
\vspace{-3mm}

\centering

\includegraphics[width=1\linewidth]{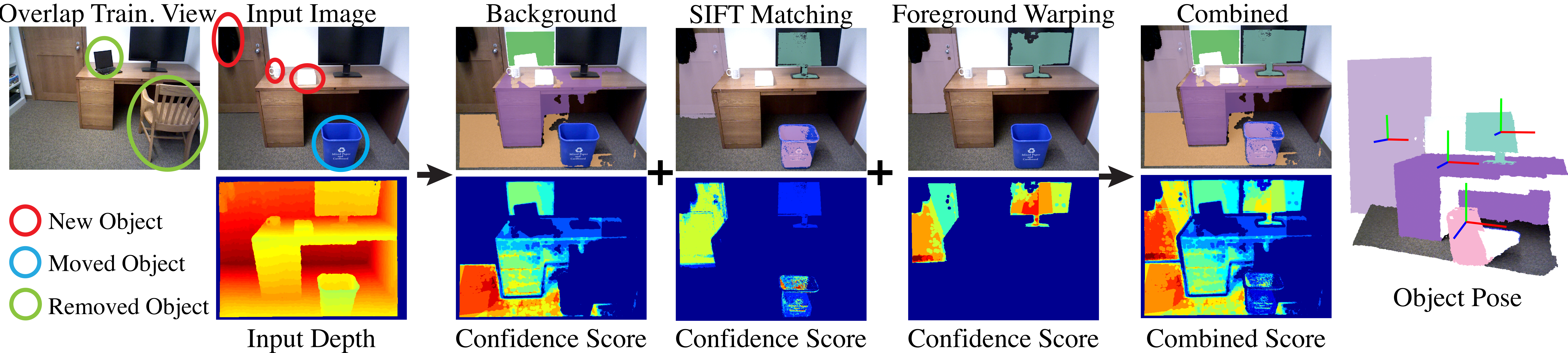}


\vspace{-1mm}
\caption{{\bf Recognition pipeline during testing.} Details are explained in Section \ref{sec:localization} and \ref{sec:objectmodel}. }
\label{fig:IntermediateResult}

\vspace{-4mm}
\end{figure*}

\vspace{-5mm}\paragraph{Instance association.}

For the each selected view, 
we ask human workers on AMT to label all the objects in this frame, 
using a LabelMe-style interface \cite{LabelMe}.
We require a instance level segmentation for each frame. 
But associating instances across many different images 
is a very challenging and time consuming task, especially when annotators work collectively.
Therefore,
we propose an automatic algorithm to associate the polygons to objects instances, utilizing our good 3D reconstruction.
For each pair of images with overlapping views,
we try to establish association between the two sets of polygons.
Using their relative pose and depth maps, 
we warp the label from one image to another,
to obtain the ID mapping pair between the two label masks at each pixel in the common overlapping area.
We count the frequencies for each matched label pairs if their object names are the same.
We use a greedy algorithm to go through all pairs one by one in the descending order of  their frequencies, until it cannot find more pairs.
For each pair, we accept it only if its frequency is more than 50 pixels, and none of the two polygons has been paired to others so far.
In this way,
from each of two images, we can get a partial association list. 
We merge all the lists into one, and use connected component to obtain the final object association for the whole sequence.
We evaluate the instance association by the ``office'' sequence. Figure \ref{fig:association} shows some associated polygons across images.
It makes no mistake of merging two objects, but it is slightly over-fragment and cannot associate 10 polygons that it should be able to.
Figure \ref{fig:3DreconResult} and \ref{fig:ViewSelection} show some point clouds colored with instance association.

\vspace{-5mm}\paragraph{Label propagation.}
Because only a very small portion of frames are labeled by humans,
and we will have more invariance if more views of an object are seen,
we propagate the label to other frames, making use of our 3D reconstruction.
For an unlabeled image, we warp\footnote{We triangulate the mesh and render them in OpenGL.} the depth, color, and label maps of the annotated images to the target image.
For each pixel, if the depth and color are consistent, we cast a vote to the label.
We accumulate votes from all annotated images and take the majority to obtain a label (Figure \ref{fig:propagate2view}).

\vspace{-2mm}

\section{Testing phase}
\label{sec:recognition}

\vspace{-1mm}

During testing, given a frame with four RGB-D images from the sensors,
the task is to estimate the pose of the robot, recognize all the previously seen objects,
and identify any unseen new objects in these images.
Assuming spatial continuity,
we also take the robot pose from the previous frame as an input.
We classify objects as movable or non-movable\footnote{For simplicity, we assume there is no deformable objects (e.g. human).},
by the annotation of object category.

\begin{figure*}[t]
\vspace{-3mm}
\centering

\includegraphics[width=1\linewidth]{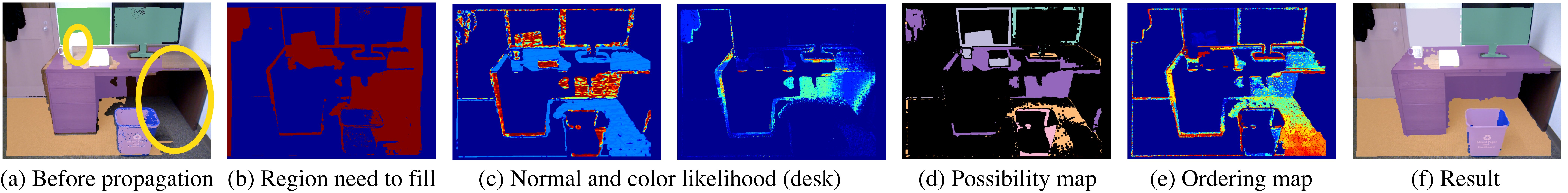}

\vspace{-1mm}
\caption{{\bf Conservative propagation.} Yellow circles in (a) highlight the holes created by the removal of occluders after the training phase.}
\label{fig:SmartProp}

\vspace{-4mm}
\end{figure*}

\vspace{-1mm}

\subsection{Localization and background subtraction}
\label{sec:localization}
\vspace{-0.5mm}

As shown in Figure \ref{fig:IntermediateResult}, 
given the initial pose,
we can identify all training frames with view overlapping.
For each of these training images,
we warp the unmovable areas to the testing view using the initial camera pose.
We check the depth consistency (with 0.1 meter as the threshold) to identify the unmovable area in the testing frame.
If the area is bigger than 50 pixels, 
we run the SIFT keypoint detection in these areas and use our pairwise alignment routine (Section \ref{sec:RGBDsfm}) 
to extract correspondences between the unmovable areas of these two frames.
We collect all correspondences from all training frames, 
and use RANSAC to find a better pose for the current testing frame.

Using this refined pose, 
to extract background, 
we repeat the same steps again to warp the unmovable areas of the training frames to the testing frame. 
For each pixel in this area with consistent depth and color, 
we accumulate a vote and threshold the vote count 
to identify the unmovable areas.
In this way, we can recognize the unmovable objects as background in the testing frame.
The remaining unrecognized areas are the foreground to be recognized next. 

\vspace{-1mm}
\subsection{Object model and recognition}
\label{sec:objectmodel}
\vspace{-0.5mm}

To recognize the foreground movable objects, 
for each object,
we build a 3D point cloud model with SIFT descriptors, similar to the state-of-the-art RGB-D instance recognition systems (e.g. \cite{AbbeelBestPaper}).
For each object,
we maintain a mixture of clusters,
each of which is a 3D point cloud merged from multiple RGB-D images, with a SIFT descriptor attached to each 3D point extracted from the image. 
We do not directly use camera pose from reconstruction to merge keypoints, 
because the reconstruction may not be perfectly accurate locally, especially for small objects.
Starting from the SIFT keypoints of the object from each frame,
we agglomeratively merge two point clouds, 
if there are enough SIFT inliers from RANSAC.
The hierarchical merging is done very conservatively,
so that we can obtain accurate 3D shapes with camera poses overfitted to each object.
Each object may be represented as a mixture of clusters, instead of only one cluster.
During testing time, for the remaining foreground areas, 
we extract SIFT keypoints with 3D locations, and use RANSAC to match these 3D points with SIFT descriptors in each cluster.
If a good matching is found, not only the category and instance label, but also the 6D object pose (Figure \ref{fig:IntermediateResult} right) are recognized.

Although movable objects may move,
they may also stay at the same location as it was during training.
Therefore, we reuse the same algorithm for background subtraction to warp the movable objects using the refined pose, 
and check consistency for both depth and color to confirm the guess.
We also accumulate a voting count to the object instance label.
This step is run in parallel with the object model matching, and the voting count is accumulated together.
Finally, we choose the label with maximal count for each pixel.

\subsection{Conservative propagation \& new object}

As shown in Figure \ref{fig:SmartProp}(b),
there may be areas that couldn't be recognized or with wrong boundaries, 
due to various small imperfection from human annotation or the algorithm.
More significantly,
there may be holes in the recognition result
because there was another object occluding this object in the training set (the yellow circles in Figure \ref{fig:SmartProp}(a)).
Now that the occluder moves away, 
a new part of the object appears, and we desire to the algorithm to be able to propagate the label to these missing areas.
However, this propagation must be very conservative,
because we do not want 
the label to propagate to cover any new object (e.g. the white box in Figure \ref{fig:SmartProp}(f)).
The correct label for new objects should be ``no label'', so that the system can send it to ask for annotation from crowd sourcing.

To handle wrong labels near the boundaries,
we first shrink the recognition mask by image erosion (with a 5-pixel disk as kernel),
and then propagate the labels to better align with the image and depth boundary.%
\footnote{During erosion, thin objects might completely disappear and not be able to recover during propagation.
So we use image opening to identify these thin objects and only shrink the labels for the thick objects.}
We treat the shrunk labels as hard constraints during label propagation.

For the propagation,
we first identify a list of possible object instance labels for each unrecognized pixel (Figure \ref{fig:SmartProp}(d)), by considering color, normal, and 3D bounding box of an object.
For each object, we can obtain their color and surface normal distributions,
and use these two distributions to estimate the likelihood of each pixel belongs to this object (Figure \ref{fig:SmartProp}(c)).%
\footnote{For a given pixel, we compute the likelihood by counting the number of pixels within the object mask with similar color or normal directions.
The count is thresholded into binary decision to indicate whether this pixel can be possibly labeled as the object.}
We also obtain a 3D bounding box for each object using the labeled 3D point cloud from the training set.
In testing, we position the bounding box using the estimated object pose,
and check whether this pixel is outside the box and therefore should not belong to this object.

\begin{table}[t]
\setlength{\tabcolsep}{2.4pt}

\footnotesize
\begin{tabular}{c|c|c|c|c|c|c}
\hline 
 & train & labeled & obj & test & labeled & area (sq.ft.)\tabularnewline
\hline 
kitchen & 432\hspace{-0.2mm}$\times$\hspace{-0.2mm}4 & 33 & 29 & 339\hspace{-0.2mm}$\times$\hspace{-0.2mm}4 & 18 & 110\tabularnewline
\hline 
office & 637\hspace{-0.2mm}$\times$\hspace{-0.2mm}4 & 58 & 40 & 415\hspace{-0.2mm}$\times$\hspace{-0.2mm}4 & 24 & 180\tabularnewline
\hline 
tea area & 21,407\hspace{-0.2mm}$\times$\hspace{-0.2mm}4 & 182 & 329& 6,779\hspace{-0.2mm}$\times$\hspace{-0.2mm}4 & 40 & 1,870\tabularnewline
\hline 
meeting place & 24,059\hspace{-0.2mm}$\times$\hspace{-0.2mm}4 & 136 & 82  & 7,001\hspace{-0.2mm}$\times$\hspace{-0.2mm}4 & 40 & 1,530\tabularnewline
\hline 
\end{tabular}

\vspace{1mm}
\caption{{\bf Benchmark statistics.}
Col 1: number of training frames.
Col 2: number of training frames being labeled.
Col 3: number of object instance labeled in the training set.
Col 4: number of testing frames.
Col 5: number of labeled testing frames with ground truth for evaluation.
Col 6: the area of the space. 
}
\label{tab:benchmark}
\vspace{-3mm}
\end{table}

\begin{figure*}[t]

\vspace{-3mm}

\def \myH {0.106}
\def \thisfolder {third_floor_tearoom_test_f126_d2}
{\small~~~~~~~Image~~~~~~~~~~~~Depth~~~~~~~~~~~~~~~NN pose~~~~~~~~\cite{Xiaofeng} pre-train ~~~\cite{Xiaofeng} re-train~~~~~background~~~~~~ before prop.~~~~ after prop.~~~~~with human}

\includegraphics[width=\myH\linewidth]{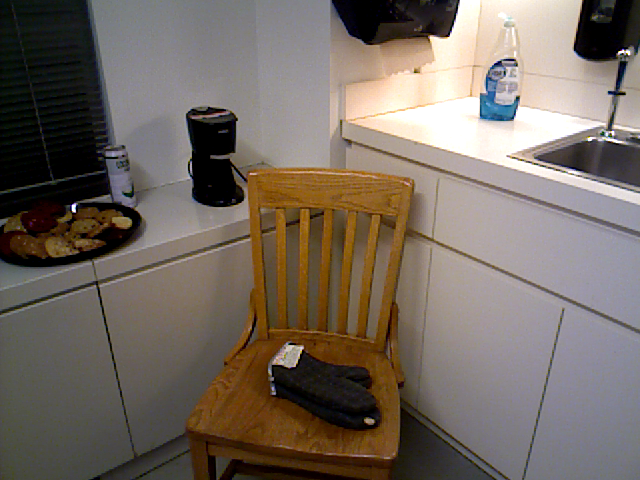}~%
\includegraphics[width=\myH\linewidth]{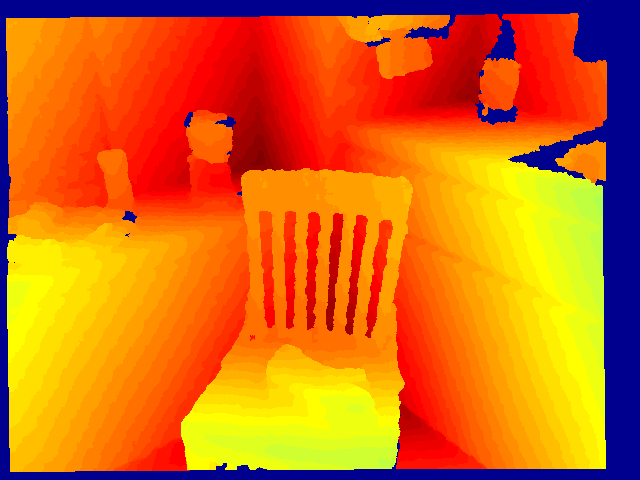}~%
\includegraphics[width=\myH\linewidth]{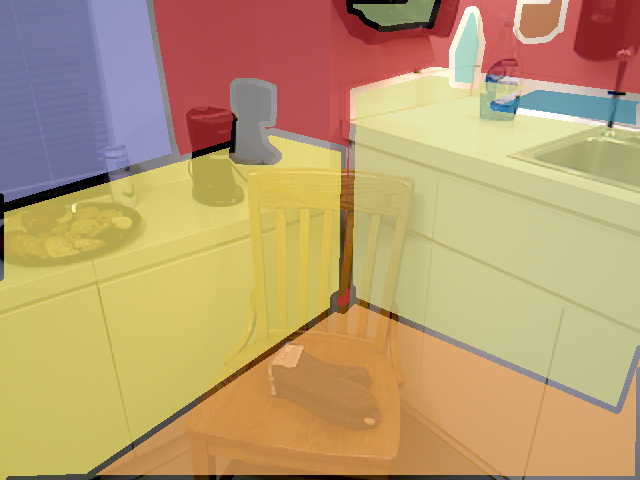}~%
\includegraphics[width=\myH\linewidth]{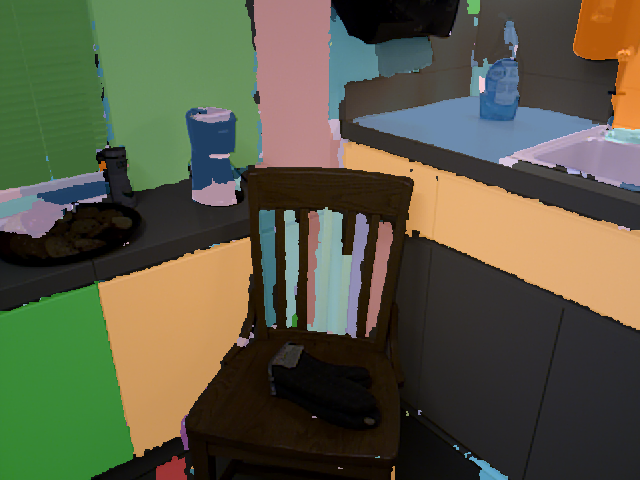}~%
\includegraphics[width=\myH\linewidth]{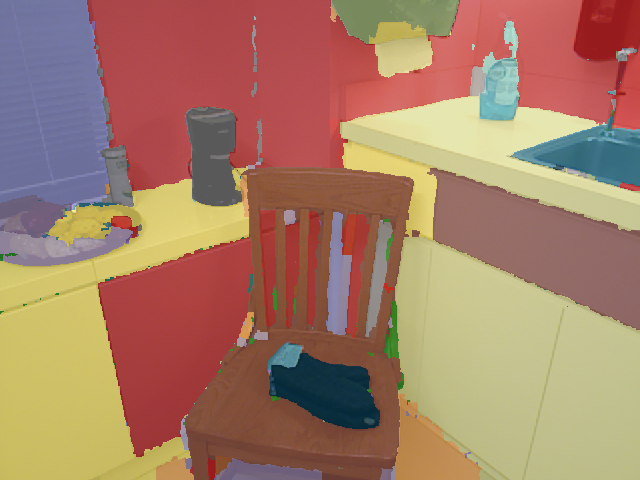}~%
\includegraphics[width=\myH\linewidth]{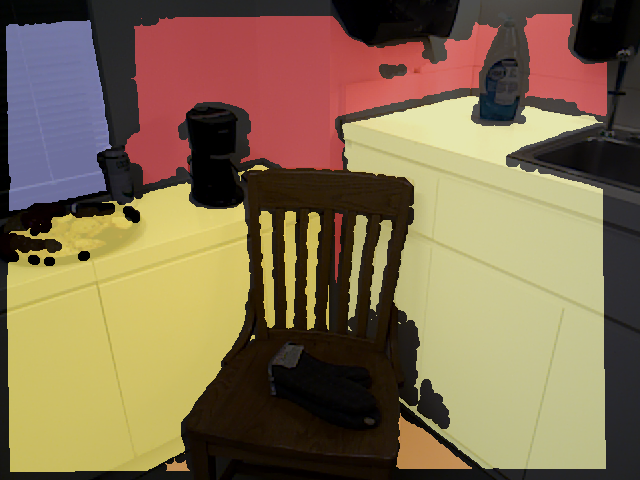}~%
\includegraphics[width=\myH\linewidth]{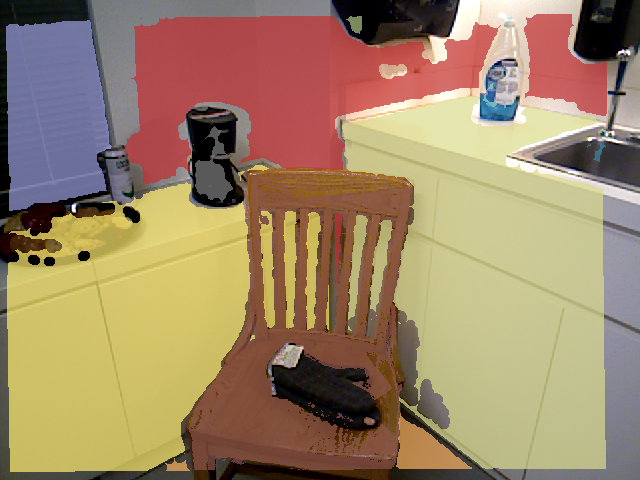}~%
\includegraphics[width=\myH\linewidth]{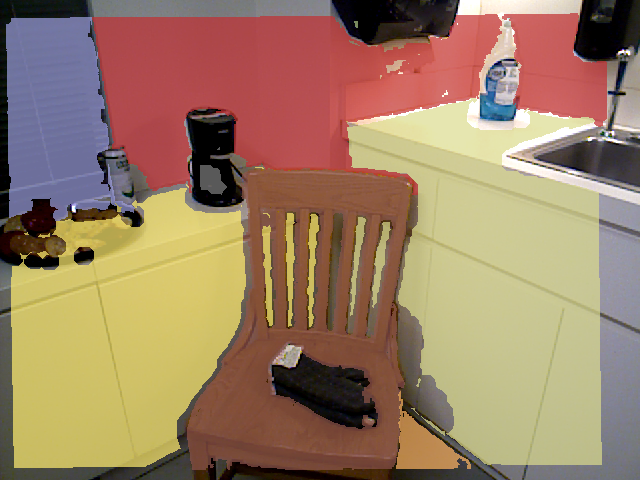}~%
\includegraphics[width=\myH\linewidth]{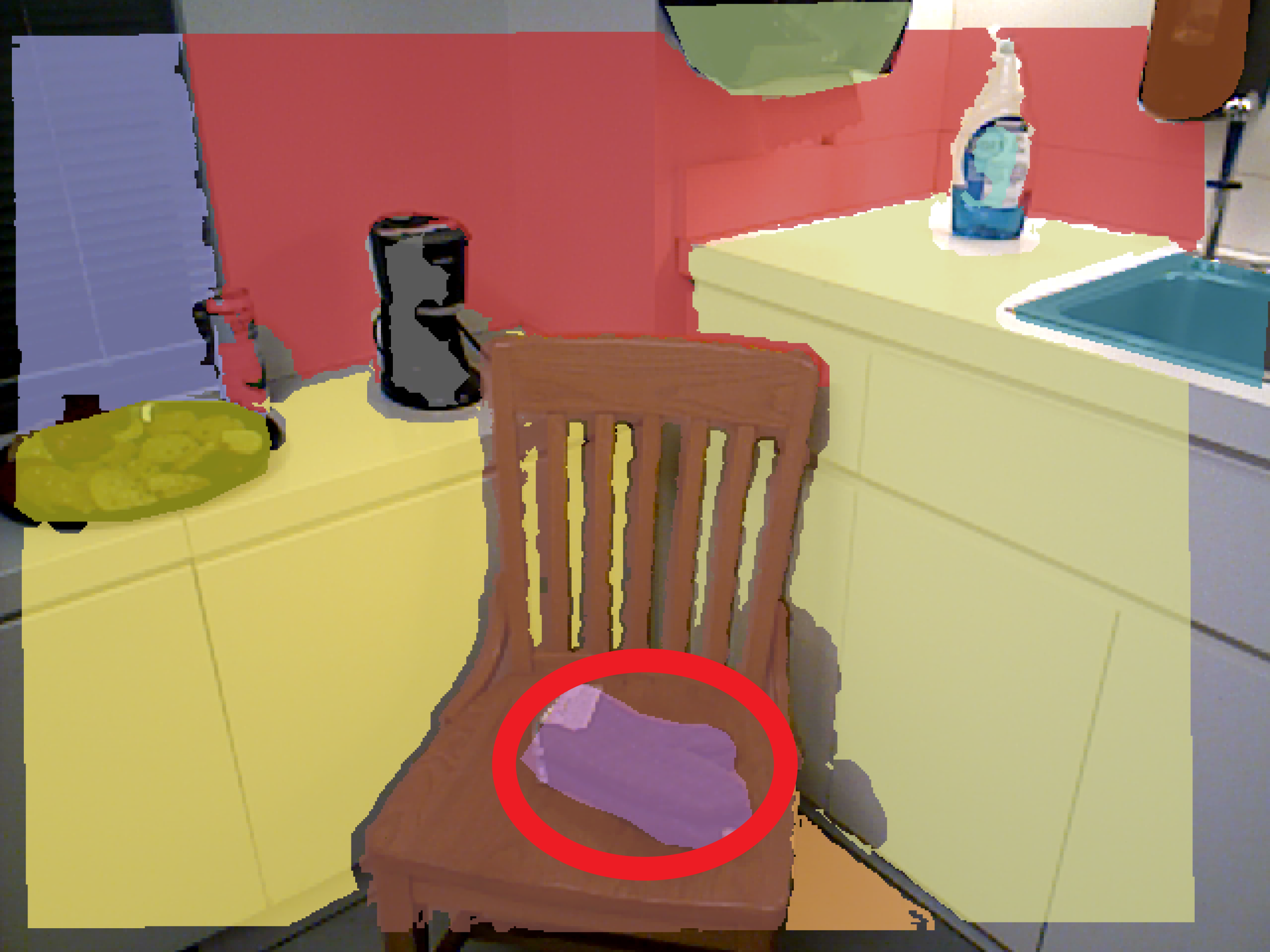}
\vspace{0.7mm}

\def \thisfolder {fields_center2_testing_f1_d2}
\includegraphics[width=\myH\linewidth]{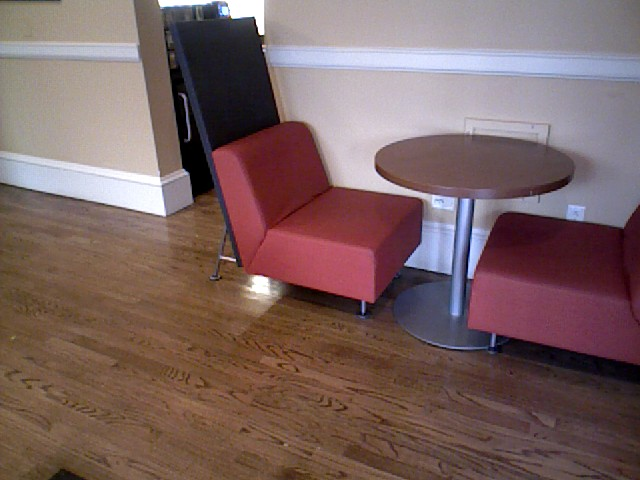}~%
\includegraphics[width=\myH\linewidth]{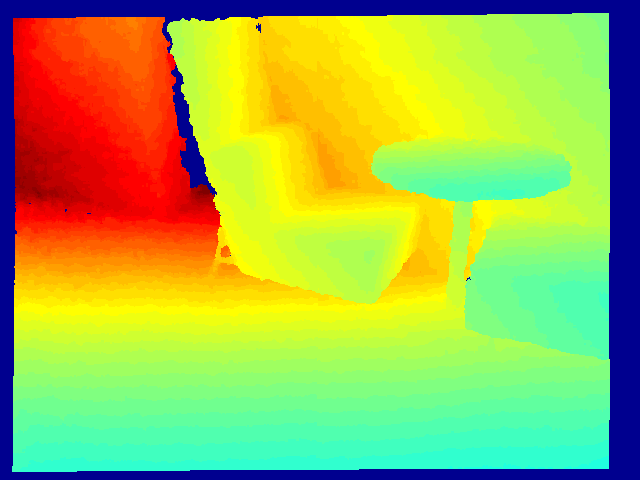}~%
\includegraphics[width=\myH\linewidth]{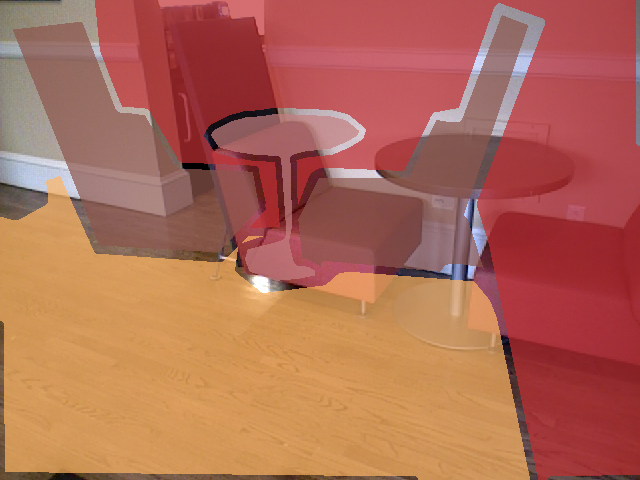}~%
\includegraphics[width=\myH\linewidth]{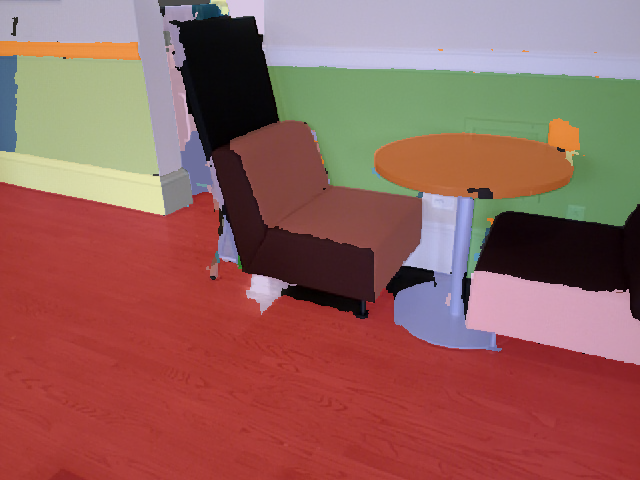}~%
\includegraphics[width=\myH\linewidth]{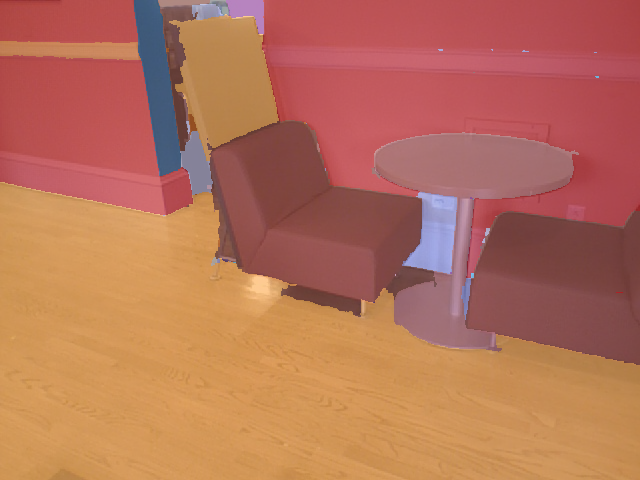}~%
\includegraphics[width=\myH\linewidth]{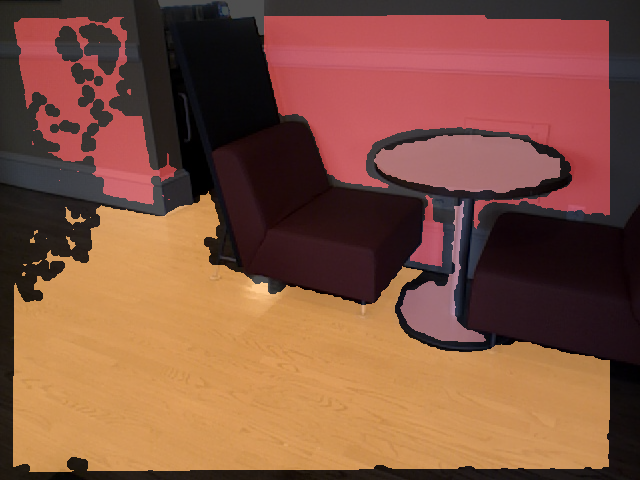}~%
\includegraphics[width=\myH\linewidth]{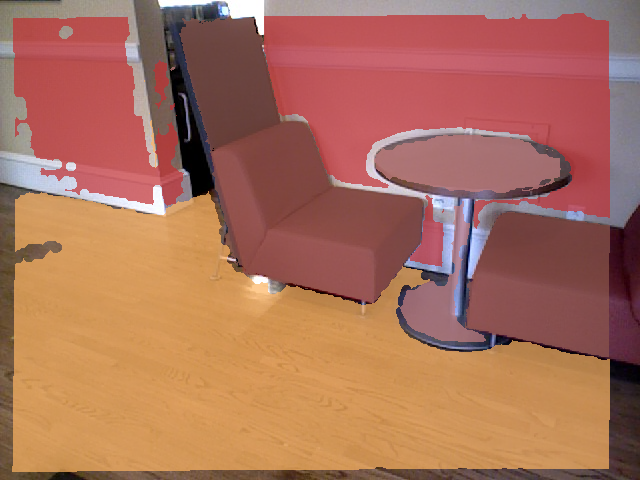}~%
\includegraphics[width=\myH\linewidth]{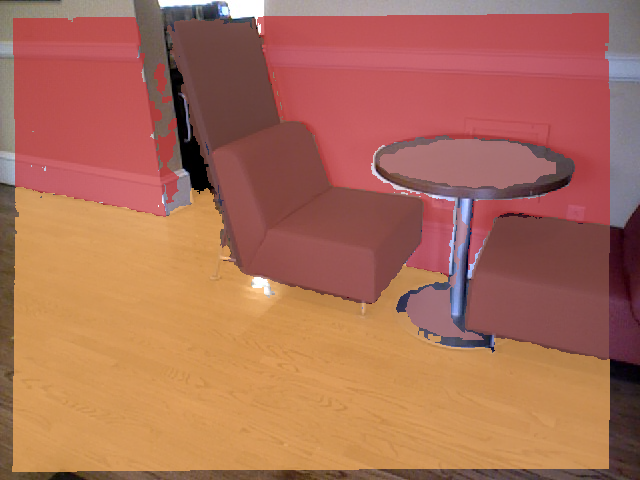}~%
\includegraphics[width=\myH\linewidth]{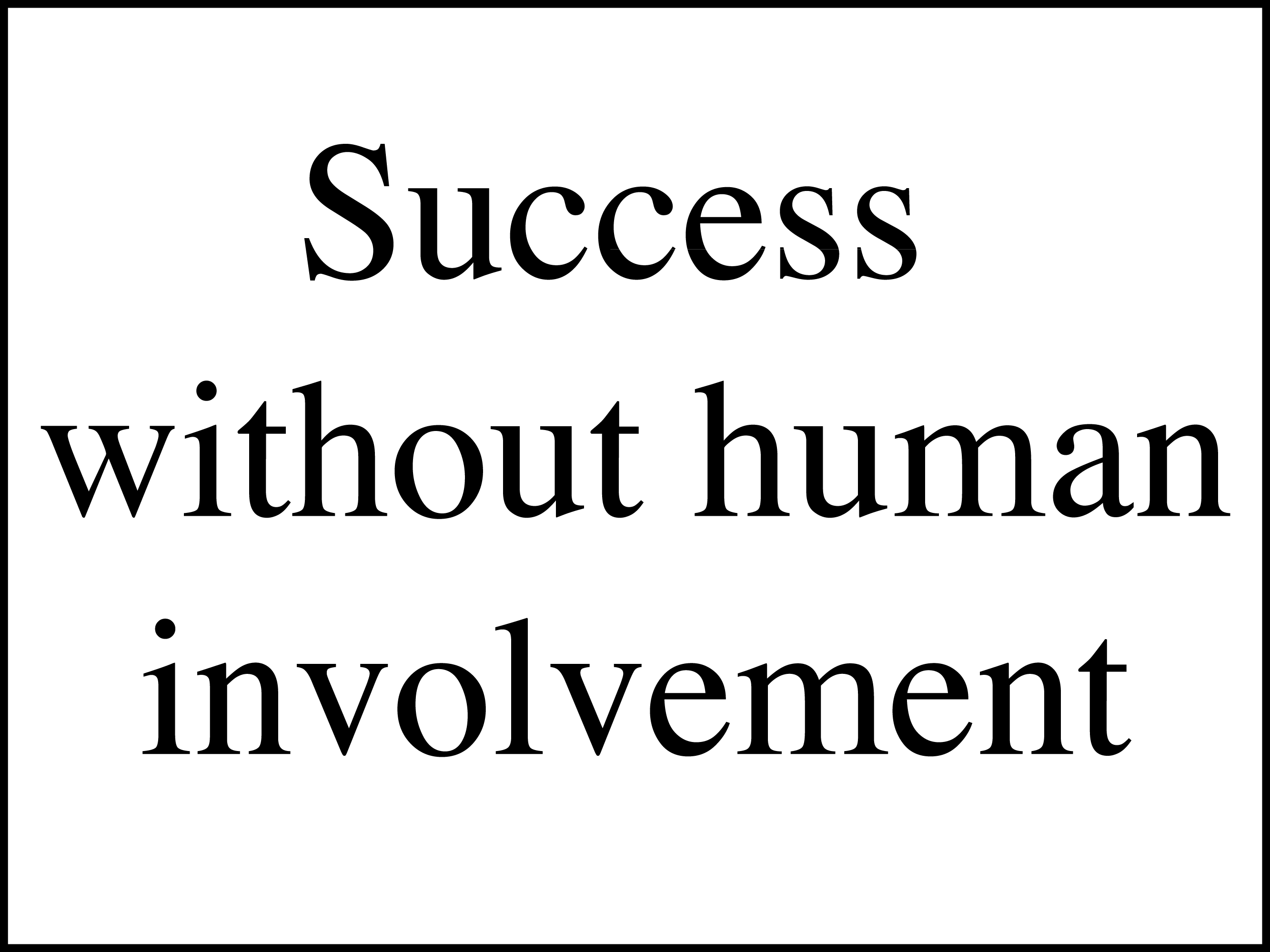}
\vspace{0.7mm}

\def \thisfolder {third_floor_tearoom_test_f26_d2}
\includegraphics[width=\myH\linewidth]{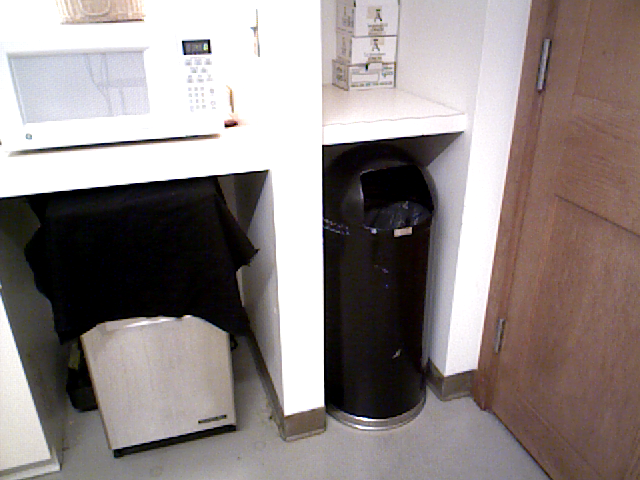}~%
\includegraphics[width=\myH\linewidth]{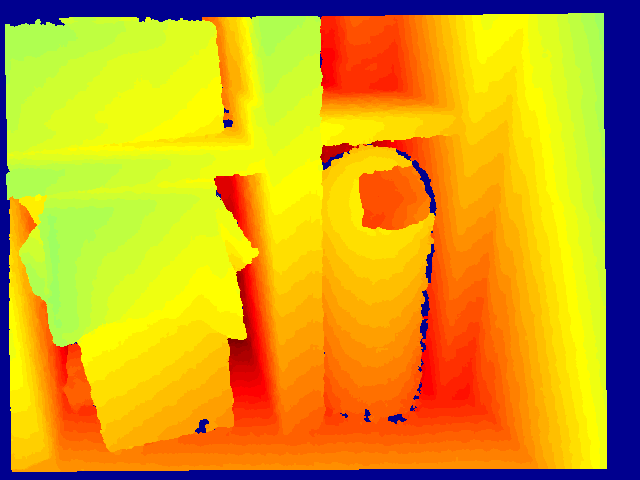}~%
\includegraphics[width=\myH\linewidth]{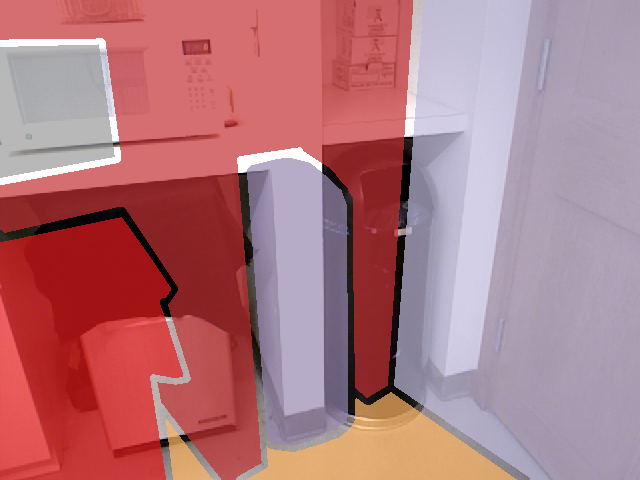}~%
\includegraphics[width=\myH\linewidth]{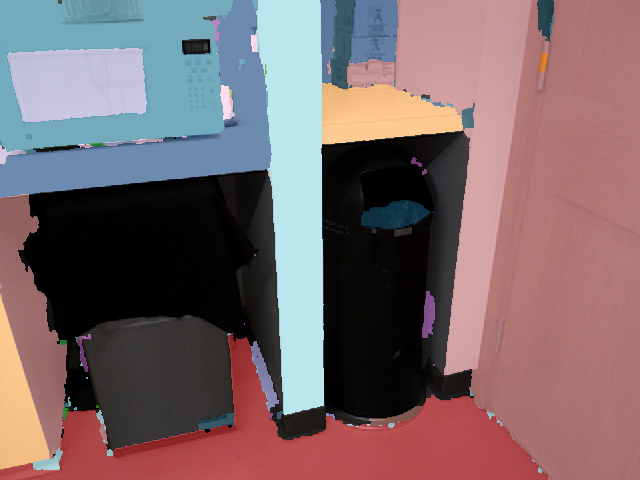}~%
\includegraphics[width=\myH\linewidth]{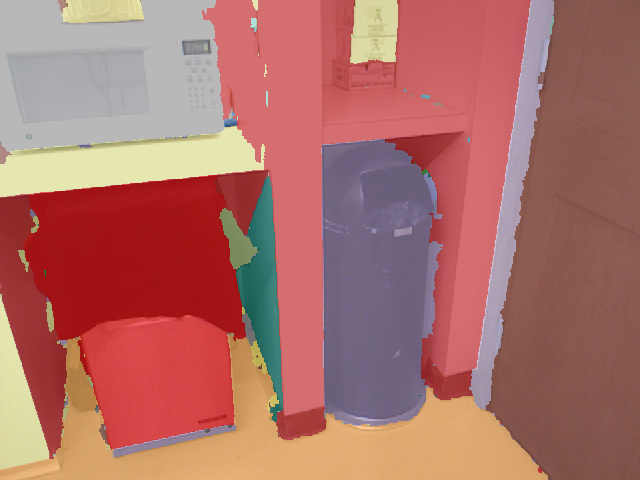}~%
\includegraphics[width=\myH\linewidth]{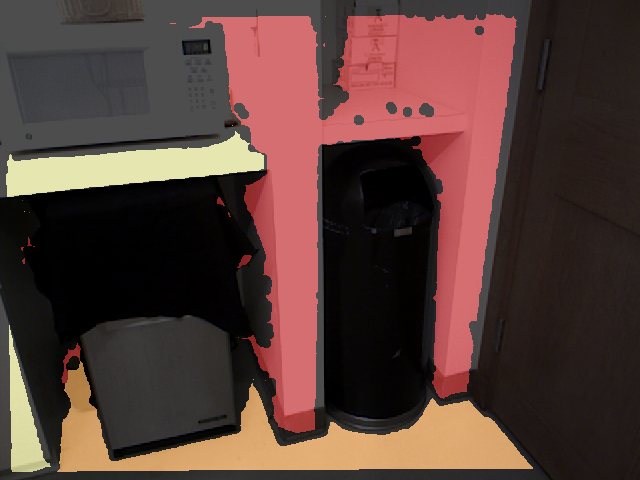}~%
\includegraphics[width=\myH\linewidth]{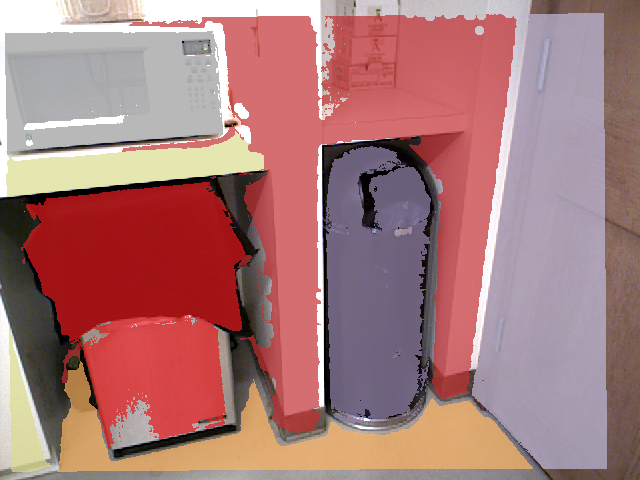}~%
\includegraphics[width=\myH\linewidth]{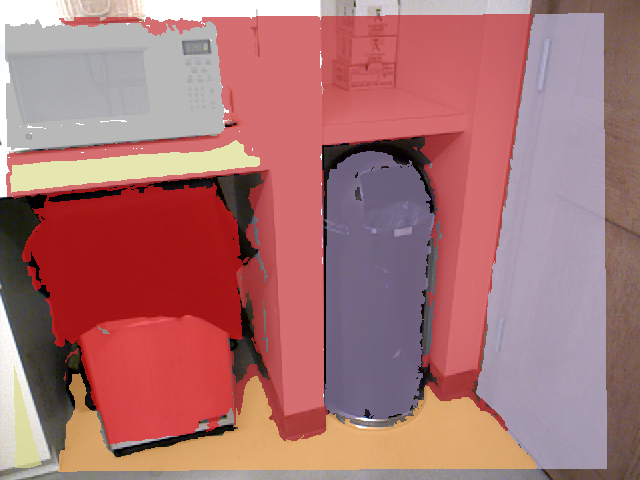}~%
\includegraphics[width=\myH\linewidth]{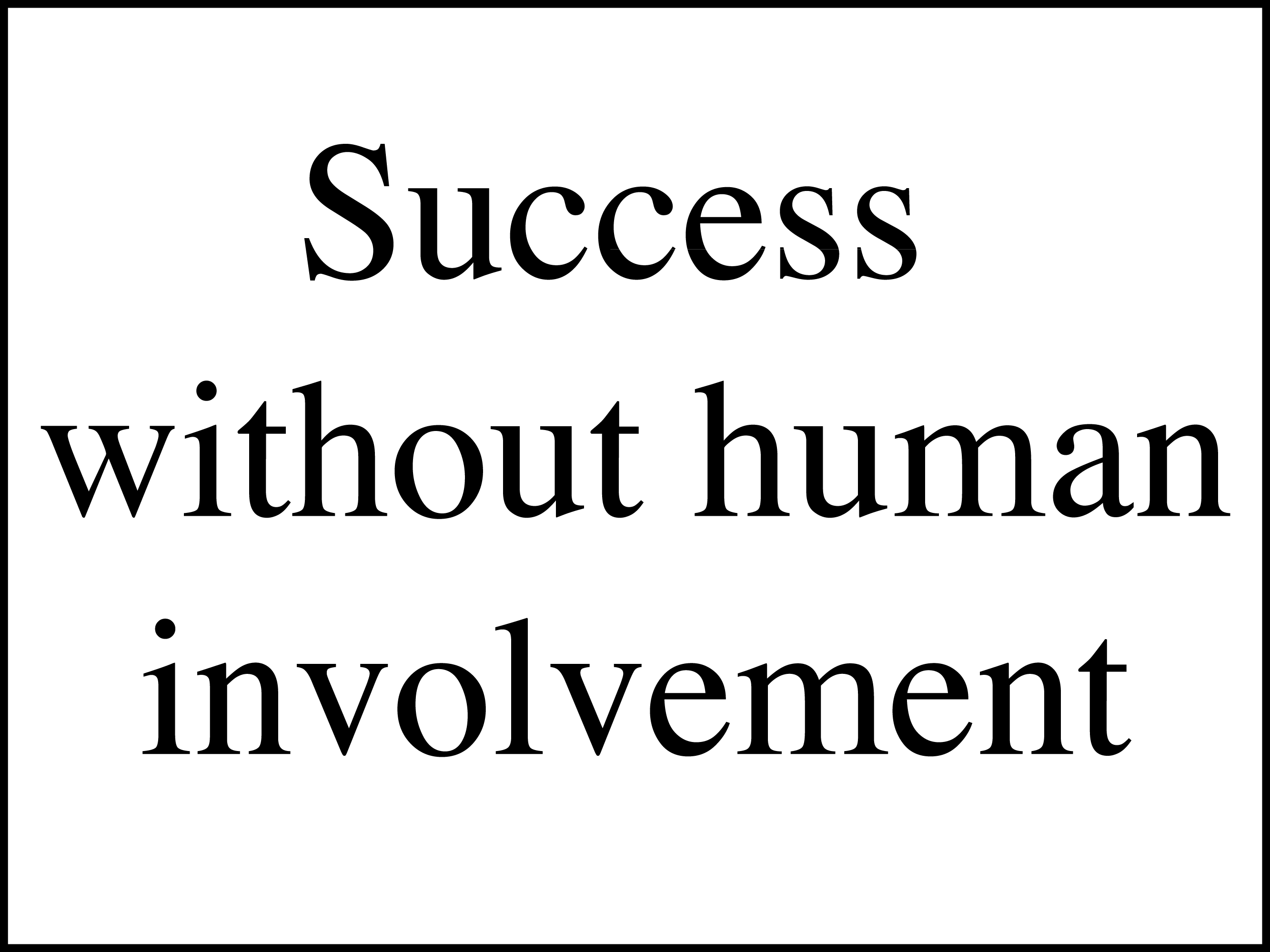}
\vspace{0.7mm}

\def \thisfolder {jx_office_testing_f451_d3}
\includegraphics[width=\myH\linewidth]{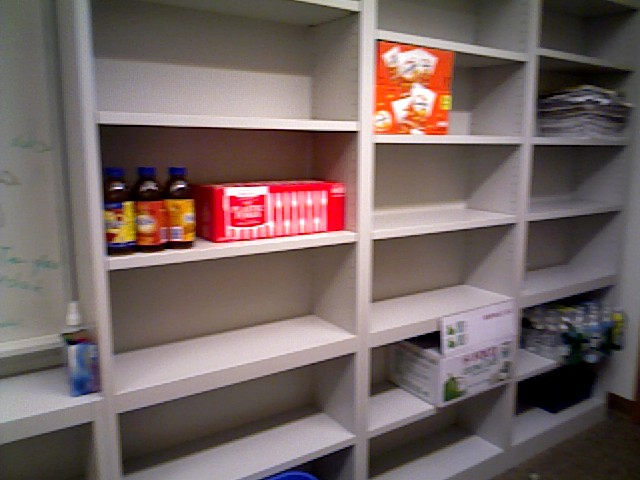}~%
\includegraphics[width=\myH\linewidth]{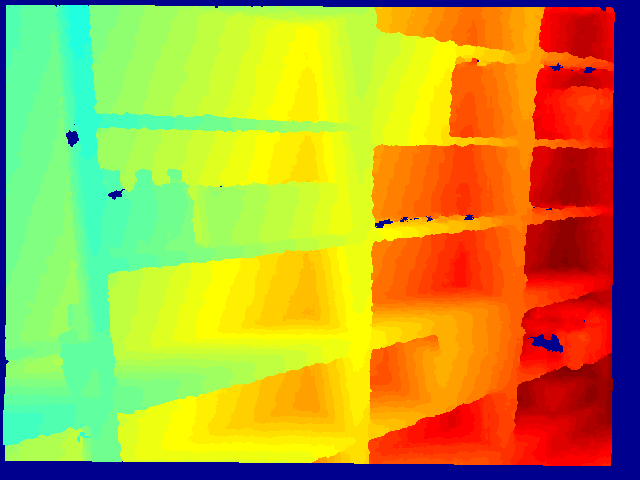}~%
\includegraphics[width=\myH\linewidth]{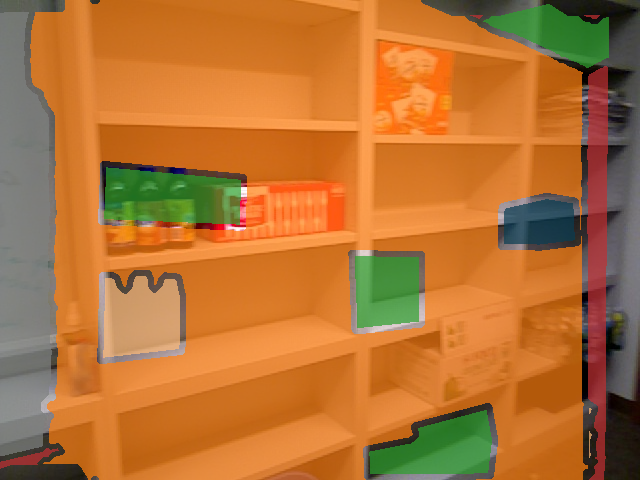}~%
\includegraphics[width=\myH\linewidth]{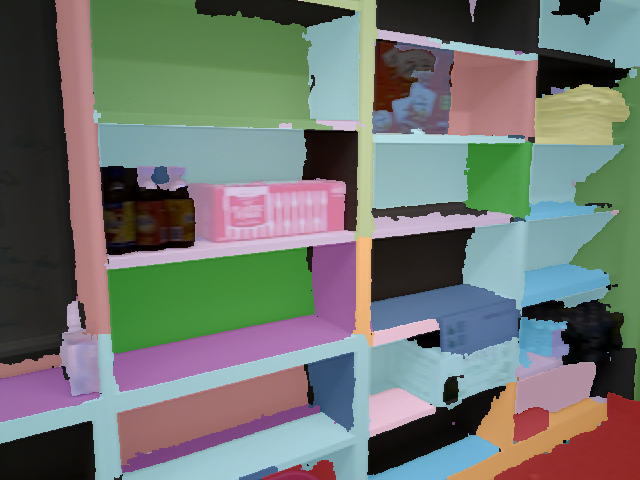}~%
\includegraphics[width=\myH\linewidth]{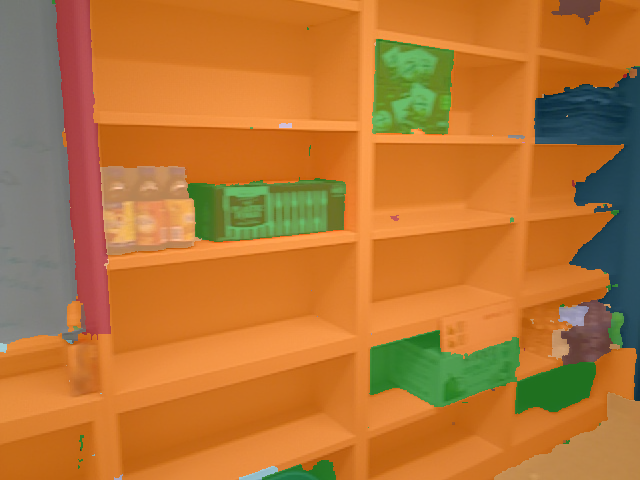}~%
\includegraphics[width=\myH\linewidth]{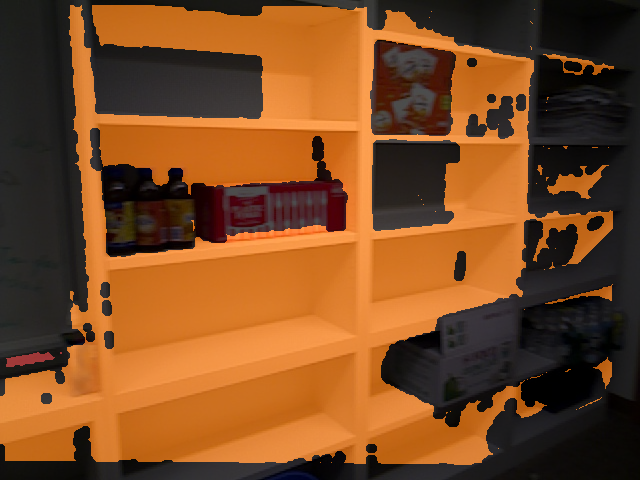}~%
\includegraphics[width=\myH\linewidth]{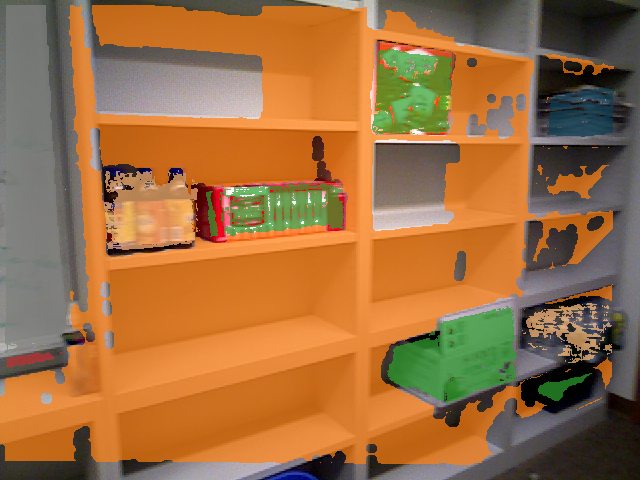}~%
\includegraphics[width=\myH\linewidth]{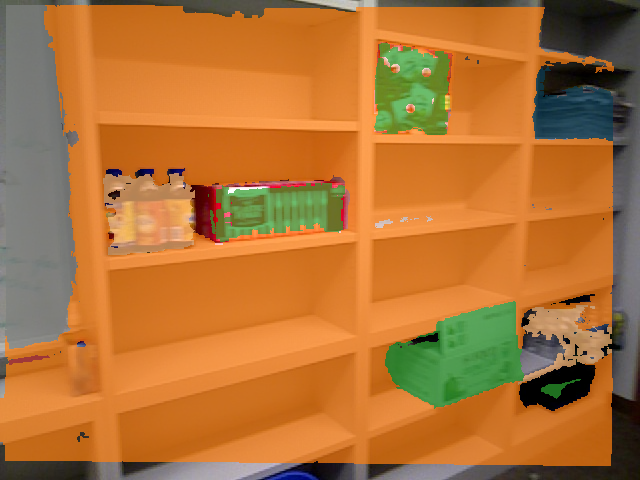}~%
\includegraphics[width=\myH\linewidth]{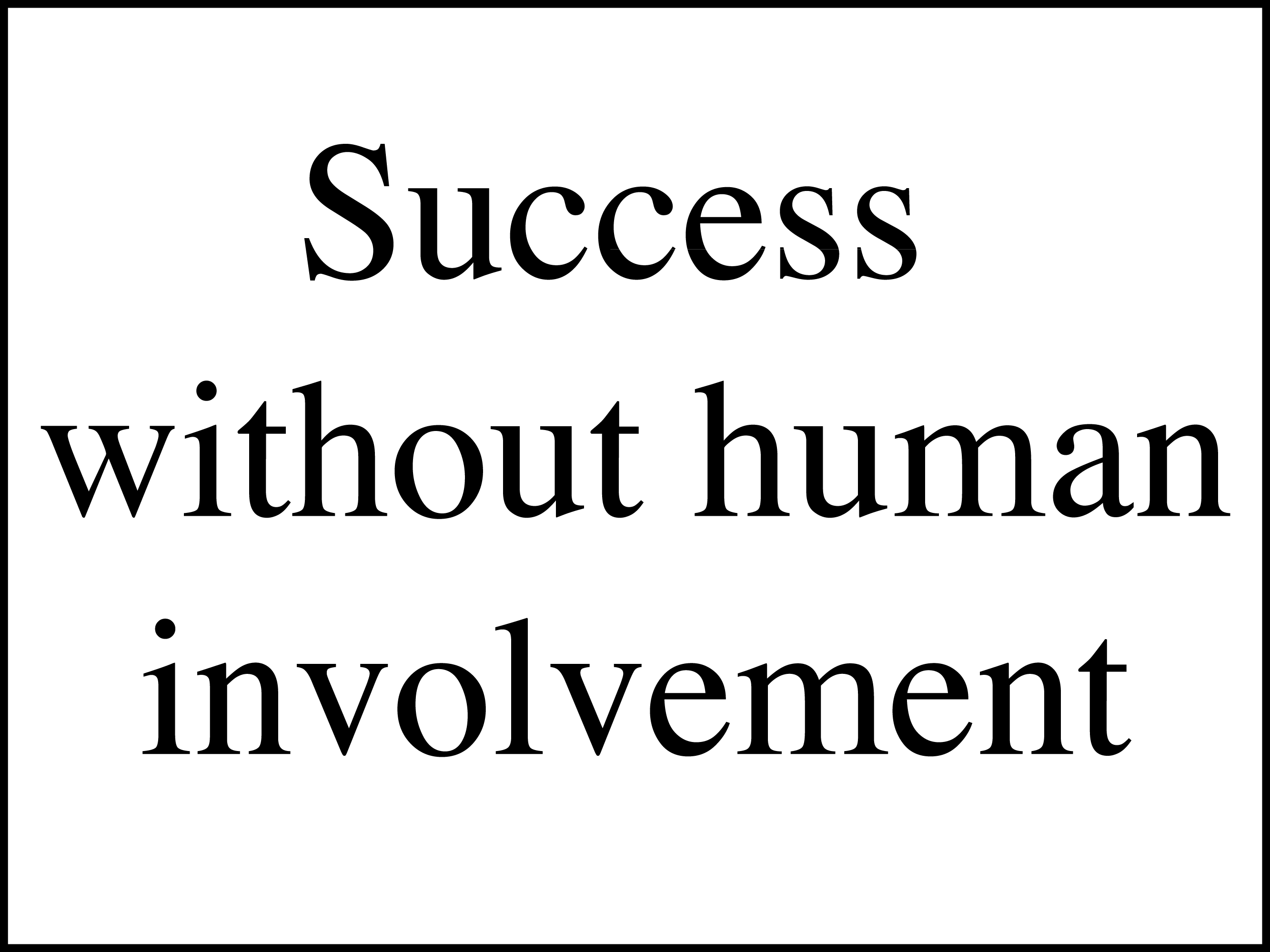}
\vspace{0.7mm}

\def \thisfolder {jx_office_testing_f451_d1}
\includegraphics[width=\myH\linewidth]{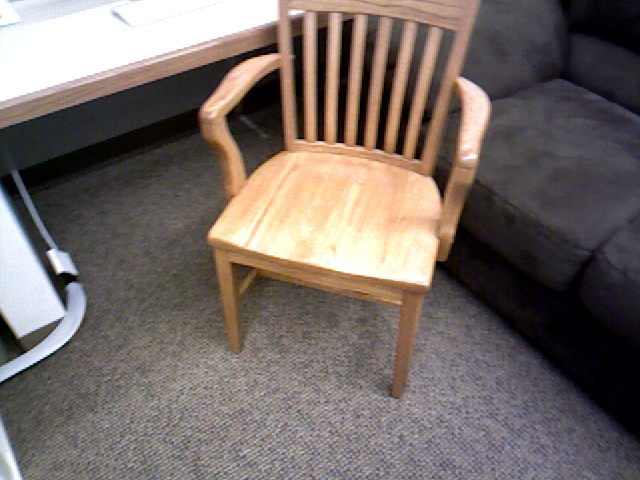}~%
\includegraphics[width=\myH\linewidth]{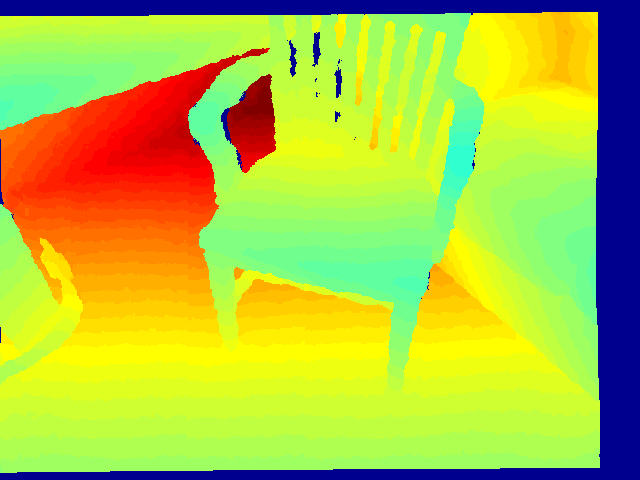}~%
\includegraphics[width=\myH\linewidth]{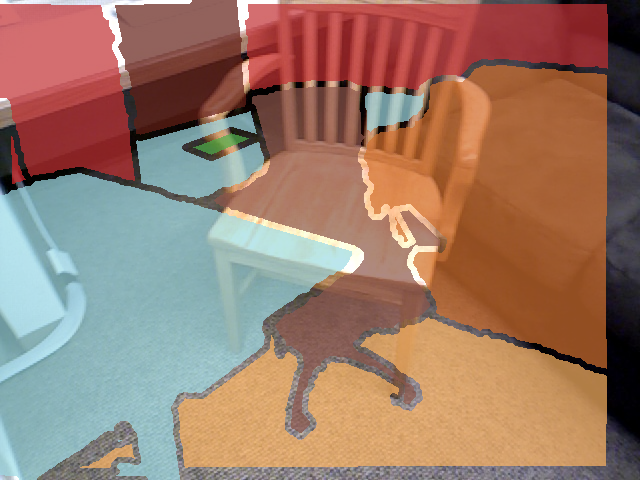}~%
\includegraphics[width=\myH\linewidth]{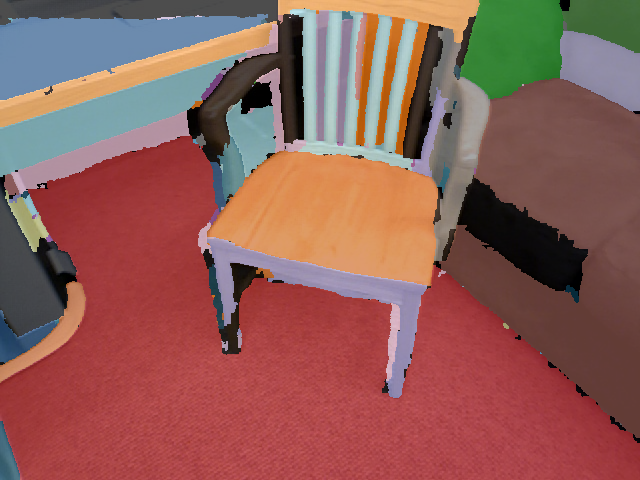}~%
\includegraphics[width=\myH\linewidth]{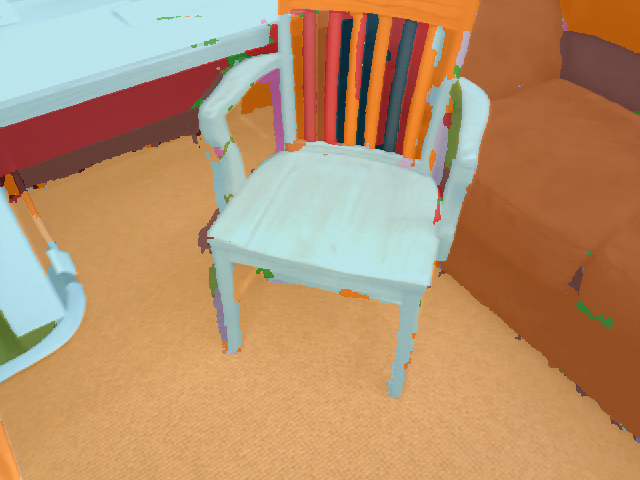}~%
\includegraphics[width=\myH\linewidth]{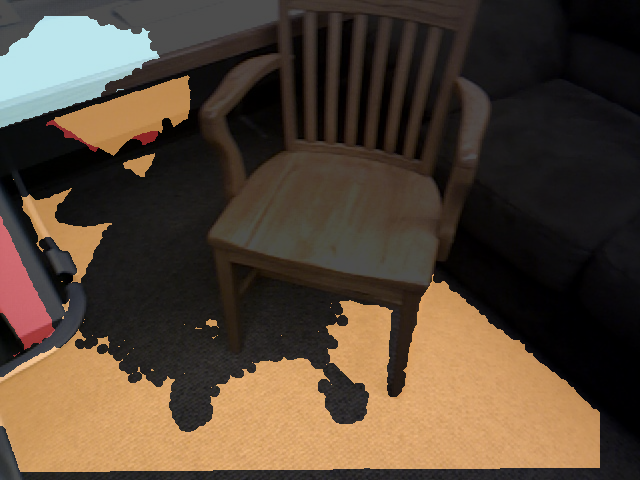}~%
\includegraphics[width=\myH\linewidth]{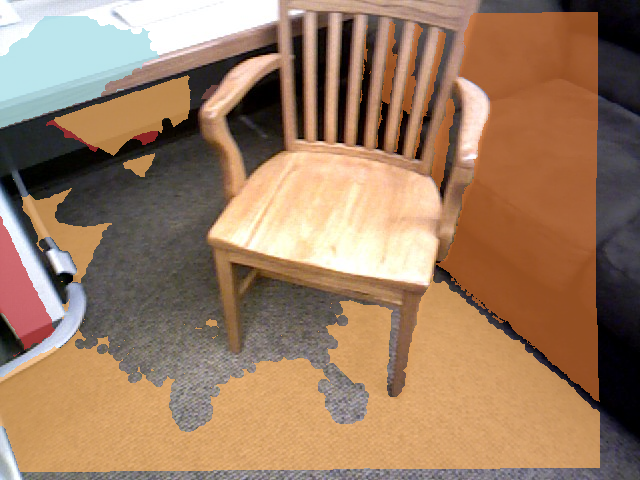}~%
\includegraphics[width=\myH\linewidth]{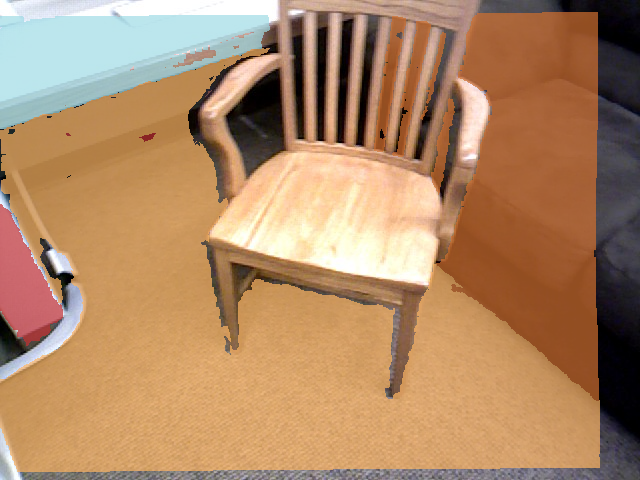}~%
\includegraphics[width=\myH\linewidth]{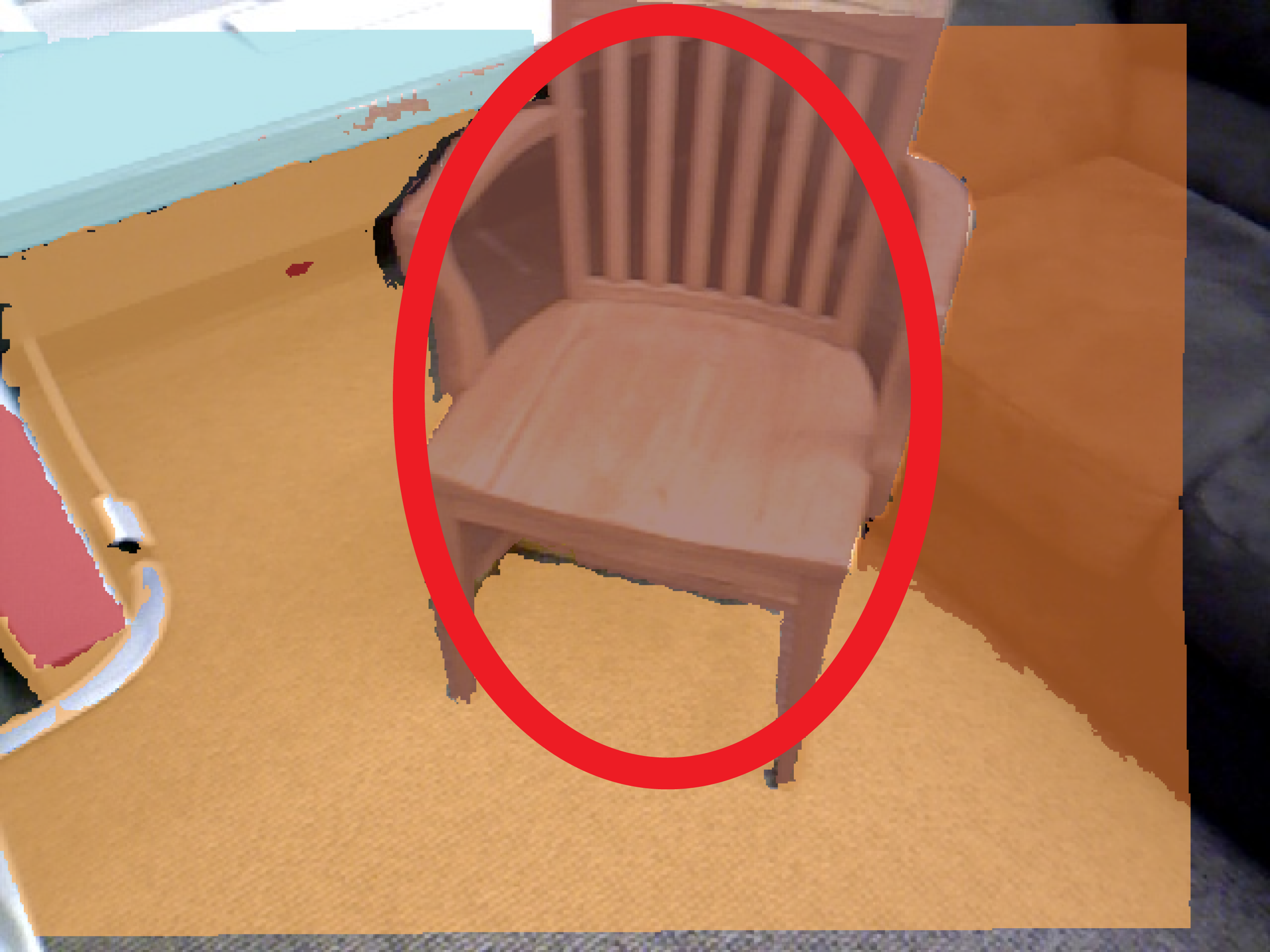}


\vspace{-0.8mm}

\includegraphics[width=1\linewidth]{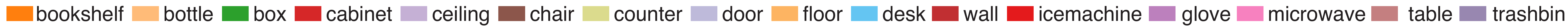}

\vspace{-1mm}

\caption{{\bf Result comparison.} Here we show the category-level semantic segmentation result. Red circles highlight the new objects.
}
\label{fig:VisualizeComparison}

\vspace{-2mm}
\end{figure*}
\begin{table*}[t]
\footnotesize
\setlength{\tabcolsep}{3pt}
\begin{tabular}{c|c|c|c|c|c|c|c|c|c|c|c|c}
\hline 
 &  & \cite{Xiaofeng} pre-train & \cite{Xiaofeng} re-train & NN gist & NN pose & bg+fgwarp & bg+SIFT & SIFT & bg+fgwarp+SIFT & + propagate & +human & no new\tabularnewline
\hline 
\multirow{2}{*}{instance} & recall & - & 84.45 & 62.12 & 63.95 & 70.15 & 60.72 & 67.48 & 74.01 & 86.58 & 91.01 & 86.59\tabularnewline
\cline{2-13} 
 & precision & - & 84.45 & 66.67 & 68.56 & 97.61 & 97.97 & 96.82 & 97.44 & 96.03 & 96.24 & 96.23\tabularnewline
\hline 
\multirow{2}{*}{category} & recall & 6.67 & 86.24 & 62.12 & 63.95 & 70.15 & 60.72 & 67.48 & 74.01 & 86.58 & 91.01 & 87.36\tabularnewline
\cline{2-13} 
 & precision & 6.67 & 86.24 & 67.96 & 69.85 & 98.5 & 98.45 & 97.95 & 98.35 & 96.89 & 97.13 & 97.08\tabularnewline
\hline 
\end{tabular}

\vspace{1mm}
\caption{{\bf Performance evaluation.}}
\label{tab:eval}
\end{table*}

A good propagation should have a smooth transition of color, normal and depth.
And it should be able to stop if it is a new object.
We also desire to enforce the spatial continuity and want the propagation to be connected to the hard constrained label mask.
To this end,
we maintain a max-heap of all unlabeled pixels adjacent to the areas with labels.
In each iteration,
we will choose one pixel from the heap that has the most similar color, normal, and depth with one of its neighboring labeled pixel 
to propagate the label from this neighbor, without violating the label possibility constraints estimated above.
When a pixel is chosen to propagate the label, 
its unlabeled neighbors will be put into the heap, or their key values will be updated (i.e. increase-key) by the similarity score with this newly labeled pixel.
The iterations stop
when the maximal similarity score is below a set threshold.
Figure \ref{fig:SmartProp}(e) shows the pixel ordering of the propagation.
In such a way, we make use of color, normal, depth, size, continuity to propagate conservatively, 
and stop when new object(s) are introduced (Figure \ref{fig:SmartProp}(f)).

After propagation,
if there are still a large unrecognizable area,
the image will be sent to crowd sourcing platform to request for annotation. 
Otherwise,
the recognition is considered successful autonomously.
In both cases, the label mask is integrated to the object model. 
The whole pipeline for testing takes about 10 mins per frame for a typical scene.

\vspace{-1mm}
\section{Experiments}
For this new task, 
we construct a benchmark to evaluate the algorithm carefully.
The benchmark also enables offline comparison without a robot, 
and eases follow-up research.

\vspace{-1mm}


\subsection{Benchmark evaluation}
\vspace{-1mm}

We construct a benchmark of four places: an office, a kitchen, a tea area, and a meeting place.
Table \ref{tab:benchmark} gives some basic statistics. 
The kitchen is a single room in an office building.
The office is a professor's office.
The tea area contains two connected rooms: 
a tea room  for social events with a small kitchen,
and a lounge with sofas and a coffee table.
The meeting place contains three connected rooms:  
a meeting room with a comfortable setting for discussion groups of 20-25 people,
a large living room to provide a cozy space,
and a kitchen.
For the training phase, 
the robot exhaustively scans the places using rotate-and-move scheme.
For the testing phase, 
we move some old objects naturally and introduce some new objects.
For labeling,
we experiment with AMTurk and we can obtain good results with quality control mechanisms from \cite{SUNRGBD}.
But to construct a high quality benchmark, we label all data in house. 

We evaluate the recognition result by semantic segmentation accuracy.
We use recall to measure the ratio of pixels with correct labels to the total number of pixels. 
We only evaluate on the set of pixels $\mathcal{I}$ with valid depth and annotation.
We use precision to measure the percentage of correct pixels among all the predicted labels. 
In our case,
we desire an almost perfect precision with a high recall,
so that robot can avoid requesting humans to annotate too frequently.
When a new object appears, we desire the algorithm to predict ``no label'' at the region, 
and prediction of any label will be considered as wrong. 
For the $i$-th pixel, let $t_i$ be its ground truth label, and $l_i$ be the predicted label ($l_i = \emptyset$ when it predicts no label).
Let $\mathcal{N}$ denote the set of new objects.
Therefore, the correctness of a predicted label is $\delta_i = (t_i\notin \mathcal{N} \land l_i = t_i) \lor (t_i\in \mathcal{N} \land l_i = \emptyset)$.
And the recall is $ \frac{\sum_{i\in \mathcal{I}}{\delta_i}}{\left|\mathcal{I}\right|}$, and the precision is
$ \frac{\sum_{i\in \mathcal{I}}{\delta_i}}{\sum_{i\in \mathcal{I}}{(t_i\notin \mathcal{N} \land l_i \neq \emptyset) \lor (t_i\in \mathcal{N} \land l_i = \emptyset)}}$.


\vspace{-1mm}
\subsection{Comparisons}
\vspace{-1mm}

To see the benefits from constraining a robot to stay in a closed environment,
in Table \ref{tab:eval}, we compare the performances of an state-of-the-art algorithm \cite{Xiaofeng} pre-trained on SUN-RGBD \cite{SUNRGBD} vs. re-trained on our training set.
The bad performance of the pre-trained model shows that general object recognition is very difficult.
And the huge performance gap between them suggests that constraining in a closed environment makes the problem much easier. 
Different from our method, this approach mostly relies on appearance, 
without any other cues such as object movability, environment map and camera pose. 
Note that this approach \cite{Xiaofeng} 
has the same recall and precision, because it makes prediction on every pixel regardless of its confidence, which is undesirable in our scenario.
We also design another set of baselines using Nearest Neighbor (NN), based on appearance or camera pose. 
For appearance-based NN, we extract GIST features on RGB-D image \cite{SUNRGBD} to find the NN.
For camera-pose-based NN, we use the initial pose to find the training frame that has the largest overlapping view.
In both cases, we directly copy the NN's label map as the result.
Table \ref{tab:eval} and Figure \ref{fig:VisualizeComparison} shows the comparison with our algorithm.
By using all domain cues, our algorithm significantly outperforms all these baselines.

To evaluate the importance of each component in our algorithm, we decompose our pipeline and evaluate them separately.
``bg+fgwarp'' assumes none of the objects moves and it simply warps the label from training frames using refined camera poses. 
``bg+SIFT'' uses warping to explain the non-movable objects and uses SIFT to match all the remaining regions. 
``SIFT'' assumes that all objects are movable and directly matches them with object models using SIFT.
``bg+fgwarp+SIFT'' combines all above the components. 
``+ propagate'' is our full pipeline that also propagates the label to increase prediction coverage. 
The results in Table \ref{tab:eval} suggest that every component in our pipeline is very reliable with a very high precision.
Our final result combines different components to increase the coverage of the prediction, in order to achieve a better recall. 
``+human'' shows what happens if the algorithm decides to ask a human annotator to fix the result using crowdsourcing.
We can see that human performance is not 100\%, which is caused by the imperfection of annotation on boundaries. 
Last but not least, 
``no new'' shows the performance when there is no new object (we exclude the new object area in evaluation). 
From here, we can see that the automatic algorithm performs almost as good as humans.
This suggests that when there is no new object, the level of accuracy is high enough for fully autonomous operation without any human involvement.

\vspace{-2mm}

\section{Conclusion}

\vspace{-1mm}

We observe that most robots will stay in a closed environment (e.g. a house).
Based on this, we propose a way to enable a robot to recognize all objects at an almost perfect accuracy,
leveraging 3D maps and crowd sourcing.
By formulating and constructing a benchmark,
we hope to lay the foundation of a new direction in vision for robotics.

\paragraph{Acknowledgement.}
This work is supported by Princeton Project X grant. We thank Thomas Funkhouser and for valuable discussion.
{\small
\bibliographystyle{ieee}
\bibliography{robot,pvg}
}
\end{document}